\documentclass[letterpaper]{article} % DO NOT CHANGE THIS
\usepackage{aaai25}  % DO NOT CHANGE THIS
\usepackage{times}  % DO NOT CHANGE THIS
\usepackage{helvet}  % DO NOT CHANGE THIS
\usepackage{courier}  % DO NOT CHANGE THIS
\usepackage[hyphens]{url}  % DO NOT CHANGE THIS
\usepackage{graphicx} % DO NOT CHANGE THIS
\usepackage{natbib}  % DO NOT CHANGE THIS AND DO NOT ADD ANY OPTIONS TO IT
\usepackage{caption} % DO NOT CHANGE THIS AND DO NOT ADD ANY OPTIONS TO IT
\usepackage{algorithm}
\usepackage{algorithmic}

\usepackage{newfloat}
\usepackage{listings}
\DeclareCaptionStyle{ruled}{labelfont=normalfont,labelsep=colon,strut=off} % DO NOT CHANGE THIS
\floatstyle{ruled}
\newfloat{listing}{tb}{lst}{}
\floatname{listing}{Listing}
\usepackage{multirow}     
\usepackage{amsmath}
\usepackage{amssymb}
\usepackage{subcaption}
\usepackage{pifont}
\usepackage{booktabs} 
\usepackage{rotating}

\title{ContextualStory: Consistent Visual Storytelling with Spatially-Enhanced and Storyline Context}
\author {
    Sixiao Zheng\textsuperscript{\rm 1,\rm 2},
    Yanwei Fu\textsuperscript{\rm 1,\rm 2}
}
\affiliations {
    \textsuperscript{\rm 1}Fudan University\\
    \textsuperscript{\rm 2}Shanghai Innovation Institute\\
    sxzheng18@fudan.edu.cn, yanweifu@fudan.edu.cn
}

\usepackage{bibentry}
\begin{document}

\maketitle

\begin{abstract}
Visual storytelling involves generating a sequence of coherent frames from a textual storyline while maintaining consistency in characters and scenes. Existing autoregressive methods, which rely on previous frame-sentence pairs, struggle with high memory usage, slow generation speeds, and limited context integration. To address these issues, we propose ContextualStory, a novel framework designed to generate coherent story frames and extend frames for visual storytelling. ContextualStory utilizes Spatially-Enhanced Temporal Attention to capture spatial and temporal dependencies, handling significant character movements effectively. Additionally, we introduce a Storyline Contextualizer to enrich context in storyline embedding, and a StoryFlow Adapter to measure scene changes between frames for guiding the model. Extensive experiments on PororoSV and FlintstonesSV datasets demonstrate that ContextualStory significantly outperforms existing SOTA methods in both story visualization and continuation. 
Code is available at https://github.com/sixiaozheng/ContextualStory.
\end{abstract}


\section{Introduction}

Recent text-to-image (T2I) models, such as SD3~\cite{esser2024scaling}, excel at generating images from text but only produce individual images independently. Although text-to-video (T2V) models like SVD~\cite{blattmann2023stable} and Sora~\cite{videoworldsimulators2024} generate coherent videos but often feature simple scene or motion changes. In contrast, this paper focuses on \textit{visual storytelling}, which comprises generating a sequence of coherent story frames from a textual storyline in \textit{story visualization} and extending an initial frame from a textual storyline in \textit{story continuation}. This task has significant potential for educational applications, such as crafting vivid, coherent comics for storybooks. The key challenge is aligning generated frames with sentences while ensuring temporal consistency in characters and scenes. Providing sufficient context is essential due to the limited information in individual sentences.

\begin{figure}[t]
    \centering
    \includegraphics[width=\linewidth]{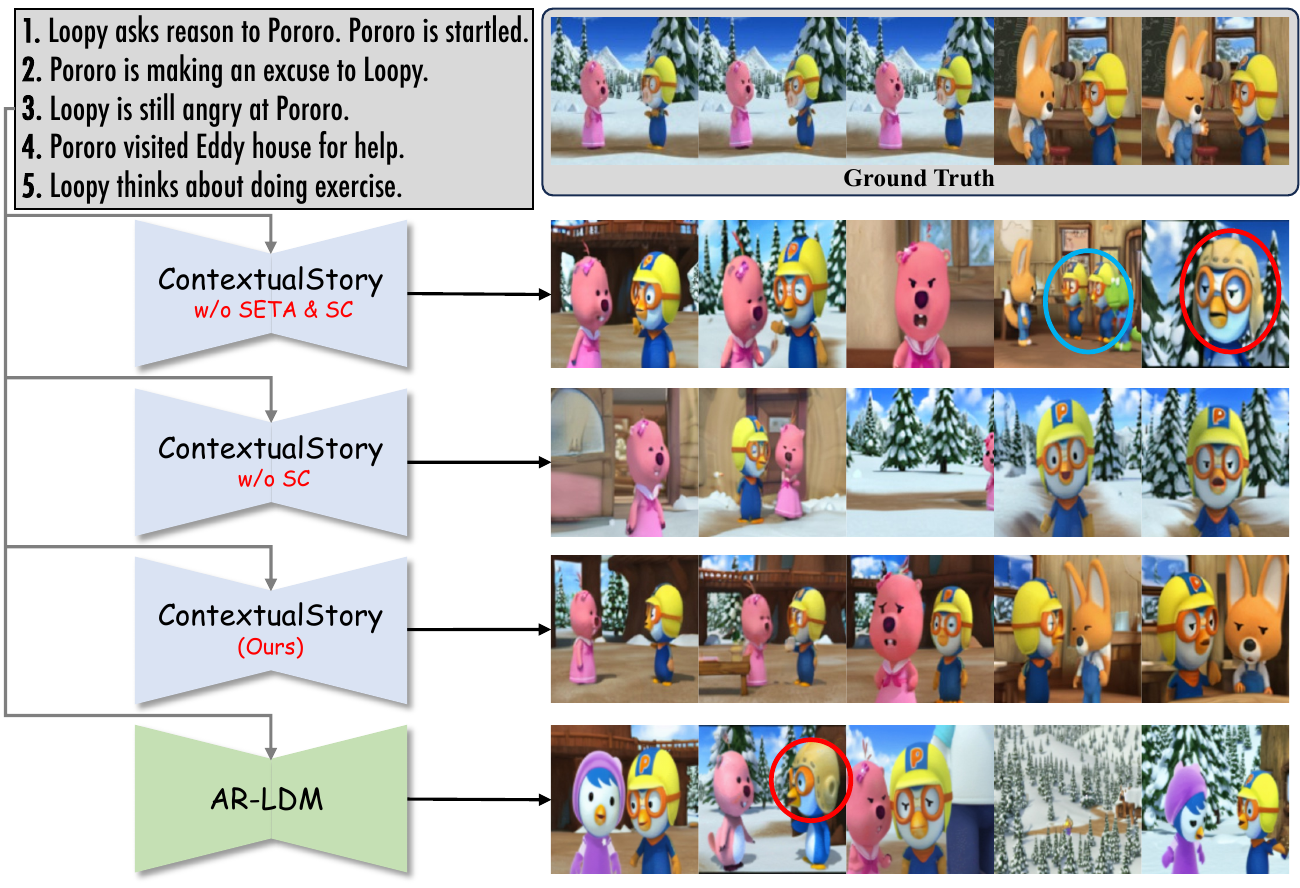}
    \caption{
    Story frames generated by our ContextualStory on PororoSV dataset. Red circles highlight character inconsistencies, and blue circles indicate repeated characters. SETA and SC enhance character consistency and scene coherence, achieving superior results compared to AR-LDM.
    }
    \label{fig:teaser}
\end{figure}

Many diffusion-based visual storytelling methods use an autoregressive generative approach to capture temporal dependencies based on previous frame-sentence pairs, such as AR-LDM~\cite{pan2024synthesizing} and Story-LDM~\cite{rahman2023make}. However, these methods face four key limitations: 1) High memory usage due to storing all previous frame-sentence pairs, making longer storyline difficult to handle; 2) Limited context in early frame generation, which may impact frame quality; 3) Slow generation speed due to the sequential nature of the process; 4) Inconsistent frames arise from relying solely on past pairs and neglecting future context, missing the global story context. We address this by exploring how the model can 1) \textit{access sufficient frame context} and 2) \textit{obtain adequate context from the storyline.}

To access sufficient frame context, we integrate temporal convolutions and Spatially-Enhanced Temporal Attention (SETA) into the UNet, combining them with the spatial modeling layer. By alternating between spatial and temporal modeling, the model  effectively captures spatial dependencies within individual frames and temporal dependencies across frames for comprehensive context. 
To obtain adequate context from the storyline, we propose the Storyline Contextualizer (SC), which processes the CLIP text embeddings to propagate the context information across sentences, providing sufficient context throughout.

Temporal attention is crucial for visual storytelling as it propagates context across frames along the temporal dimension. However, vanilla temporal attention struggles with significant \textit{character movement} between frames, as shown in Figure \ref{fig:SETA}(a). To overcome this, we propose SETA that employs a local window mechanism to allow queries to attend to features within local windows of other frames. This effectively captures moving characters and enhances spatial dependency modeling.
As shown in Figure \ref{fig:teaser}, SETA improves character consistency and reduces repeated characters, such as Pororo, compared to ContextualStory w/o SETA \& SC.

The Storyline Contextualizer enhances the contextual information of storyline embeddings from the CLIP text encoder, which initially contain only sentence-level semantics. By integrating and propagating context across sentences, it generates context-enriched storyline embeddings. The Storyline Contextualizer, a transformer-based network, incorporates self-attention and temporal attention layers to capture both global and temporal dependencies. These enriched embeddings guide the model through a temporally-aligned cross-attention mechanism, ensuring consistent story frames.
As shown in Figure \ref{fig:teaser}, ContextualStory improves scene consistency and coherence over ContextualStory w/o SC. Furthermore, compared to AR-LDM, our ContextualStory significantly enhances both characters and scenes consistency.

Additionally, to leverage scene changes between story frames to guide the model, we proposed the StoryFlow Adapter to measure change between story frames. 
For story continuation, we simply add a convolution layer at the input end of the UNet block to match the size of the first frame latent with the noise latent and then concatenate them.

Our contributions are as follows:
(1) \textbf{ContextualStory Framework}: Our novel framework overcomes limitations of existing autoregressive methods, including high memory usage, limited context, slow generation speed, and image inconsistency.
(2) \textbf{Spatially-Enhanced Temporal Attention (SETA)}: We present the SETA into the UNet model, combining temporal convolutions with spatial modeling to capture both spatial and temporal dependencies, addressing challenges of significant character movement and improving frame consistency.
(3) \textbf{Storyline Contextualizer (SC)}: A transformer-based network enriches CLIP text embeddings by capturing global and temporal dependencies, ensuring consistent story frames.
(4) \textbf{StoryFlow Adapter}: We repurpose this tool to measure scene changes between frames, guiding the model to handle scene transitions more effectively.
(5) Extensive experiments on PororoSV and FlintstonesSV datasets demonstrate that our ContextualStory significantly outperforms previous SOTA in visual storytelling.

\section{Related Works}

\noindent \textbf{Visual storytelling.}
Early methods for story visualization primarily relied on GANs~\cite{goodfellow2020generative}. 
StoryGAN~\cite{li2019storygan} pioneers story visualization using a sequential conditional GAN with a context encoder and dual discriminators to improve narrative and visual coherence. Subsequent works~\cite{song2020character,li2022clustering,maharana2021improving,maharana2021integrating,li2022word} improve StoryGAN, while others~\cite{ahn2023story,chen2022character} adopt Transformer-based methods to enhance character consistency.
StoryDALL-E~\cite{maharana2022storydall} extends the story visualization to story continuation with a given initial frame and pre-trained DALL-E~\cite{ramesh2021zero}. 
Recently, diffusion models (DM) \cite{ho2020denoising} have shown success in image generation. 
Some works \cite{pan2024synthesizing,rahman2023make,feng2023improved,song2024causal,liu2024intelligent,shen2023large,wang2024evolving} propose an autoregressive diffusion framework based on previous captions and generated frames for consistency.
For example, Story-LDM~\cite{rahman2023make} incorporates a visual memory module to capture the context of previous generated images.
However, these autoregressive methods are memory-intensive and often fail to capture the global context of the storyline. 
RCDMs~\cite{shen2024boosting} is a two-stage model that predicts the embedding of the unknown clip before generating the corresponding images.  StoryImager~\cite{tao2024storyimager} is a unified framework for story visualization, continuation, and completion. 
 StoryGPT-V~\cite{shen2023large} combines the image generation capability of LDM with the reasoning ability of Large Language Model (LLM) to ensure semantic consistency.
TaleCrafter~\cite{gong2023talecrafter}, Animate-A-Story~\cite{he2023animate}, and AutoStory~\cite{wang2023autostory} focus on designing system pipelines for story visualization, all employing LLM to generate storylines. 
In contrast, our ContextualStory addresses consistency by leveraging SETA to capture complex spatial and temporal dependencies, departing from autoregressive methods.

\begin{figure*}[ht]
    \centering
    \includegraphics[width=0.9\linewidth]{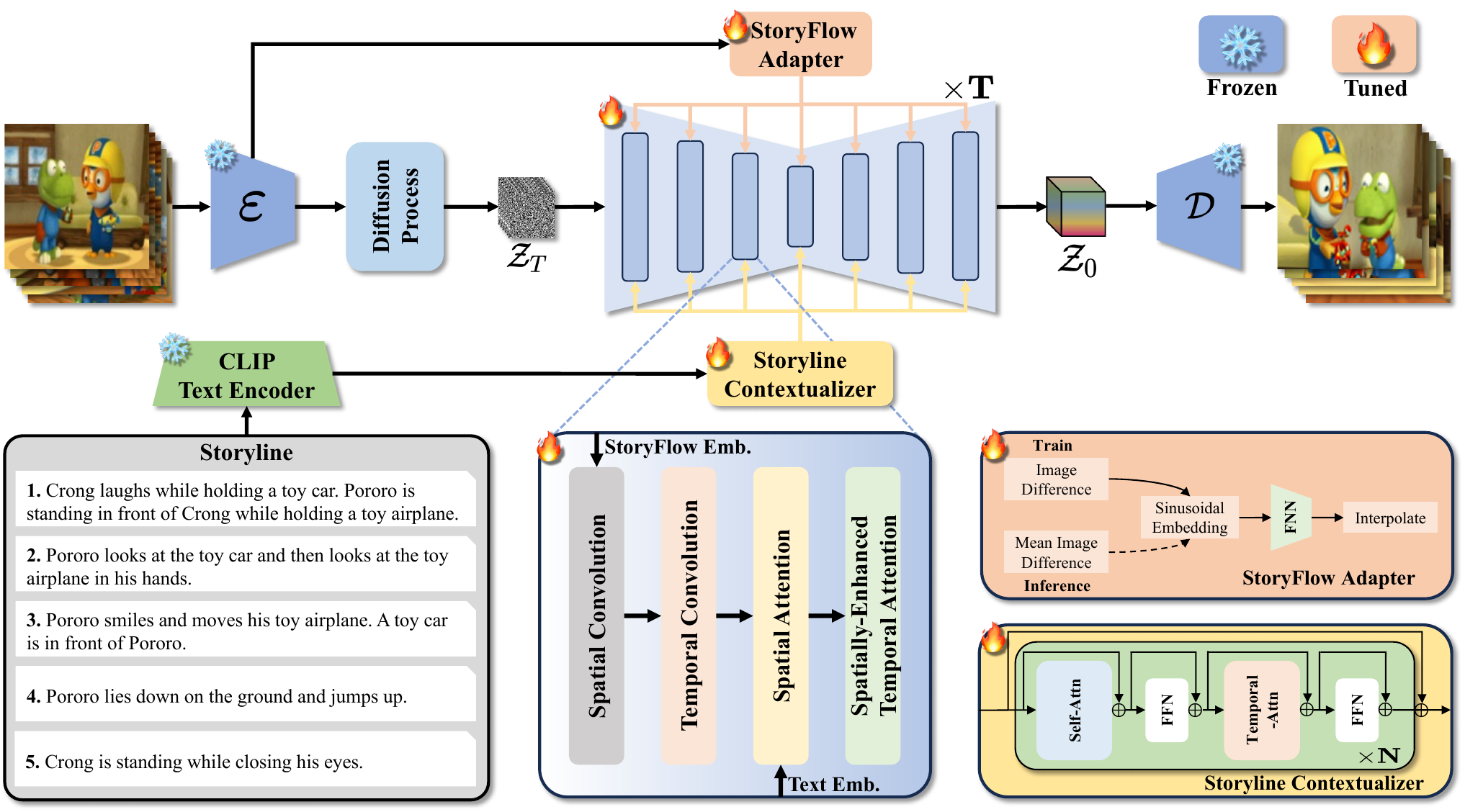}
    \caption{
    Architecture of ContextualStory for Story Visualization.
Each UNet block includes temporal convolution and \textit{Spatially-Enhanced Temporal Attention} to effectively capture complex spatial and temporal dependencies. The \textit{Storyline Contextualizer} enriches the storyline embedding by integrating context information from all text embeddings, while the \textit{StoryFlow Adapter} measures scene changes by computing differences between adjacent frames.
    }
    \label{fig:framwork_sv}
\end{figure*}

\noindent \textbf{Text-to-image generation.}
Recently, significant progress \cite{rombach2022high,saharia2022photorealistic,ramesh2022hierarchical} has been achieved in T2I generation, primarily due to advancements in DM~\cite{ho2020denoising}. 
Another line of work \cite{dhariwal2021diffusion,ho2022classifier,ruiz2023dreambooth,kumari2023multi} focuses on flexible and controllable image generation, including ControlNet \cite{zhang2023adding}, Composer \cite{huang2023composer}, IP-Adapter \cite{ye2023ip}, and T2I-Adapter \cite{mou2024t2i}. 
ControlNet provides a general pipeline for conditioning on both text and image data. 
The Diffusion Transformer \cite{peebles2023scalable} showcases scalability by replacing UNet with a Transformer, and Pixart-$\alpha$  \cite{chen2023pixartalpha} further reduces training costs while achieving superior image quality. However, these methods focus on generating individual images aligned with text and struggle to produce multiple coherent and consistent images in a sequence.

\section{Method}

Story visualization aims to generate a sequence of images $\mathcal{\tilde{I} } = \{ \tilde{I} ^1, \dots \tilde{I} ^N \}$ that align with a multi-sentence storyline $\mathcal{S} = \{ S^1, \dots S^N \}$, ensuring consistency in characters and scenes throughout. For the story continuation task, the first frame $I^1$ is provided as additional input, guiding the generation of subsequent images $\mathcal{\tilde{I} } = \{ \tilde{I} ^2, \dots \tilde{I} ^N \}$ by extracting and maintaining characters and scenes, eliminating the need to generate them from scratch. During training, ground truth images are denoted as $\mathcal{I} = \{ I^1, \dots I^N  \}$.

\subsection{Preliminaries}

Diffusion models (DM) \cite{ho2020denoising,song2020denoising} are generative models that approximate data distributions by iteratively denoising a Gaussian distribution through a reverse process of a Markov Chain. Given a training sample $\mathbf{x}_0\sim q(\mathbf{x}_0)$ and add Gaussian noise $\boldsymbol{\epsilon}\sim\mathcal{N}(\mathbf{0},\mathbf{I})$ to the input in a forward process $q(\mathbf{x}_t|\mathbf{x}_0)=\mathcal{N}(\mathbf{x}_t;\sqrt{\bar{\alpha}_t}\mathbf{x}_0,(1-\bar{\alpha}_t)\mathbf{I})$, where $\bar{\alpha}_t :={\textstyle \prod_{s=1}^t}(1-\beta_s)$ and $\beta_1,\ldots,\beta_T$ is the variance schedule. The model is trained to approximate the backward process $p_{\theta}(\mathbf{x}_{t-1}|\mathbf{x}_t)$ by minimizing the mean squared error (MSE) between the predicted and target noise  
$
    L_{DM}:=\mathbb{E}_{\mathbf{x},\boldsymbol{\epsilon}\sim\mathcal{N}(\mathbf{0},\mathbf{I}),t}\left[\|\boldsymbol{\epsilon}-\boldsymbol{\epsilon}_\theta(\mathbf{x}_t,t)\|^2\right].
$

Latent Diffusion Models (LDM) \cite{rombach2022high} extend DM to high-dimensional data by compressing images into latent space. An encoder $\mathcal{E}$ maps the input $\mathbf{x}$ to a latent representation $\mathbf{z}=\mathcal{E}(\mathbf{x})$, , where the forward and backward processes are applied. The denoising network $\epsilon_\theta(\mathbf{z}_t,t,\mathbf{c})$ is trained by minimizing 
$
    L_{LDM}:=\mathbb{E}_{\mathcal{E}(\mathbf{x}),\mathbf{c},\boldsymbol{\epsilon}\sim\mathcal{N}(\mathbf{0},\mathbf{I}),t}\left[\|\boldsymbol{\epsilon}-\boldsymbol{\epsilon}_\theta(\mathbf{z}_t,t,\mathbf{c})\|^2\right],
$
where $\mathbf{c}$ denotes conditional signals, such as storyline embeddings. The generated image $\mathbf{\hat{x}}$ is obtained by decoding the denoised latent $\mathbf{z}$ with pre-trained decoder $\mathcal{D}(\mathbf{z})$.

\subsection{Model Architecture}
\label{sec:arch_for_sv}

Previous methods based on T2I diffusion models typically use an autoregressive approach, generating story frames sequentially with each frame relying on the preceding frames and captions. However, these methods often fail to capture sufficient storyline context, leading to poor frame consistency. Moreover, the UNet struggles to capture temporal dependencies, while vanilla temporal attention layers are ineffective in addressing significant character movement across frames. To overcome these challenges, as shown in Figure \ref{fig:framwork_sv}, we introduce temporal convolution and SETA into the UNet. These components enable the model to capture contextual information across both spatial and temporal dimensions, allowing it to better handle complex spatial and temporal dependencies. We also propose the Storyline Contextualizer that ensures contextual information propagates to each sentence. Additionally, to address the significant changes in characters and scenes, we introduce the StoryFlow Adapter to quantify these changes and guide the model in generating more coherent visual stories.

\begin{figure}[t]
    \centering
    \includegraphics[width=\linewidth]{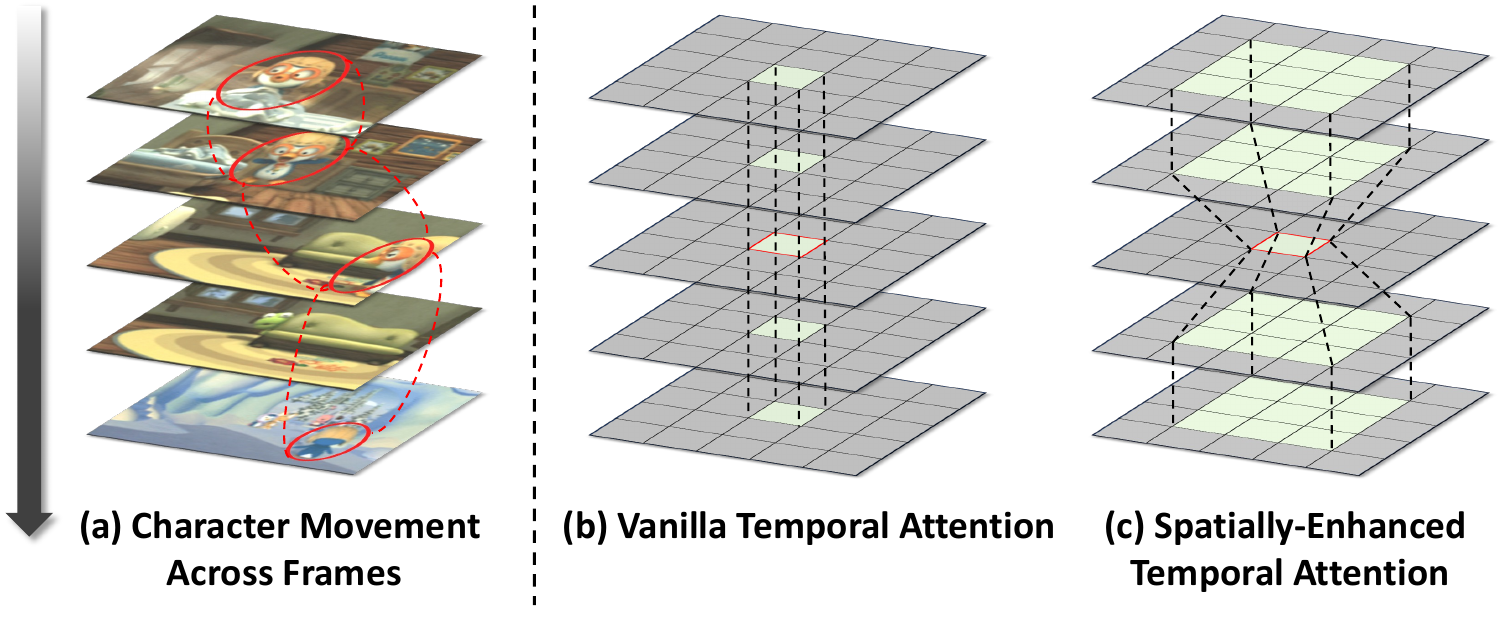}
    \caption{Spatially-Enhanced Temporal Attention leverages a local window mechanism across frames to capture both spatial and temporal dependencies, effectively handling significant character movements.
    }
    \label{fig:SETA}
\end{figure}

\noindent\textbf{Spatially-Enhanced Temporal Attention.}
In video diffusion models~\cite{zhang2023show}, temporal attention layers are often employed to model temporal dependencies. However, unlike video frames with minimal changes and redundant pixels, story frames feature significant character and scene changes. As shown in Figure \ref{fig:SETA}, significant character movement across story frames make it challenging for vanilla temporal attention to capture the same character.

To address this challenge, we propose Spatially-Enhanced Temporal Attention. Assuming the green block within the red-bordered area is the query, the query itself, along with the green blocks covered by the $k \times k$ local window at the same position across other frames (\textit{i.e.}, all the green blocks), form the key and value.
Formally, given a hidden state $\mathcal{Z}_t =\{\mathbf{z^1_t}, \dots, \mathbf{z^N_t}\} \in \mathbb{R} ^{n\times c\times h\times w}$, where $n=N$ is the number of story frames, and $c$, $h$, $w$ represent the channel, height, and width dimensions of the hidden state, respectively. 
We first reshape $\mathcal{Z} _t$ to $\mathcal{Z} _t' \in \mathbb{R} ^{hw \times n \times c}$, 
then extract the local window feature $\mathcal{Z}_t^{lw'} \in \mathbb{R} ^{h w\times n^{lw} \times c}$ at each spatial position, where $n^{lw}=(n-1)k^2+1$.
Subsequently, we compute the query, key, and value and then perform the $\mathrm{Attention} (Q_T,K_T,V_T)$ through Eq. \eqref{eq:attention}.
{\small
    \begin{gather}
    Q_S=\mathcal{Z}_t'W_S^Q ,K_S=\mathcal{Z}_t^{lw'}W_S^K,V_S=\mathcal{Z}_t^{lw'}W_S^V,\\
    \mathrm{Attention} (Q,K,V) =\mathrm{Softmax}(\frac{QK^T}{\sqrt{d} } )V,\label{eq:attention}
    \end{gather}
}
where $W_T^Q$, $W_T^K$, $W_T^V$ are learnable projection matrices, and $d$ is the feature dimensionality. To ensure a complete local window at the boundaries, we pad $\mathcal{Z} _t'$ by replicating the boundary features. We utilize rotary positional embedding (RoPE) \cite{su2024roformer} as the temporal positional embedding to enable the model understand temporal relationships between frames efficiently.

\noindent\textbf{Storyline Contextualizer.}
We use the pre-trained CLIP text encoder to independently extract text embeddings for each sentence in the storyline. These embeddings contain the semantic information of the corresponding sentences but lack the global contextual information of the storyline. Directly using these text embeddings to guide the model may result in inconsistent story frames. To address this challenge, we propose the Storyline Contextualizer, which propagates and integrates the contextual information from all text embeddings to generate a context-enriched storyline embedding. As shown in Figure \ref{fig:framwork_sv}, the Storyline Contextualizer is a transformer-based network, each layer contains a self-attention layer, a temporal attention layer, and two Feed-Forward Networks (FFNs).

Given the storyline embedding $\mathcal{C} = \{\mathbf{c^1}, \dots, \mathbf{c^N}\} \in \mathbb{R} ^{n \times l \times c_T}$ from the CLIP text encoder, where $n=N$ is the number of story sentences, $l$ is the feature sequence length, and $c_T$ is the feature dimension. 
In Storyline Contextualizer, we first reshape the storyline embedding to $1 \times nl \times c_T$ and apply self-attention, then reshape it to $l \times n \times c_T$ for temporal attention. To minimize any adverse effects from additional modules, we zero-initialize the weights of the second FFN in the final layer and incorporate a residual connection, ensuring that the Storyline Contextualizer functions as an identity mapping at the beginning of training.
The context-enriched storyline embedding $\mathcal{C}'$ guides the model through a cross-attention layer. Unlike T2I and T2V models that compute cross-attention between images/frames and a single text, we adopt a temporally-aligned approach, computing cross-attention between each individual context-enriched text embedding $\mathbf{c^i}$ and hidden state $\mathbf{z^i_t}$ of the corresponding frame.

\begin{figure}[t]
    \centering
    \includegraphics[width=\linewidth]{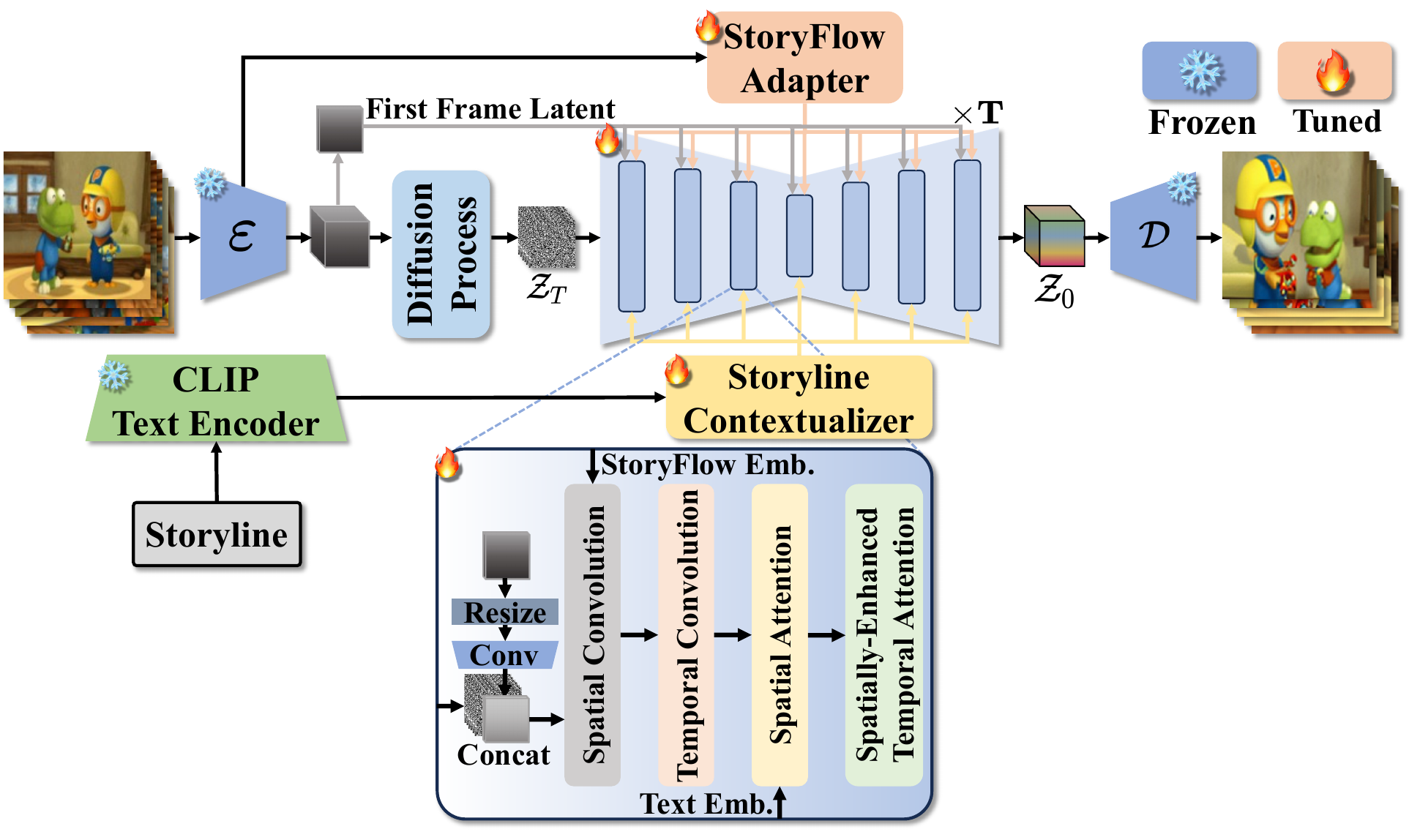}
    \caption{Architecture of ContextualStory for Story Continuation. The first frame latent is used as additional input for all UNet blocks, resized and adjusted with a $1 \times 1$ convolution layer before concatenation with the hidden state.}
    \label{fig:framwork_sc}
\end{figure}

\noindent\textbf{StoryFlow adapter.}
To leverage scene changes between story frames to guide the model, we propose the StoryFlow Adapter, inspired by \cite{jeong2023vmc, wu2023lamp, qing2023hierarchical}. As shown in Figure \ref{fig:framwork_sv}, the storyflow is computed as the L2 norm of differences between adjacent images, $\delta _i=\left\| \mathbf{z_0^i} -\mathbf{z_0^{i+1}}\right\|$, to quantify the image difference. Given $N$ ground truth frames, we calculate the storyflow $\Delta = \{ \delta ^1, \dots , \delta ^{N-1} \}$. We then encode the storyflow $\Delta$ into a $c$-dimensional embedding using sinusoidal embedding and a zero-initialized FFN. Through linear interpolation, we obtain the storyflow embedding $\Delta ' \in \mathbb{R}^{N \times c}$. Finally, we add storyflow embedding to timestep embedding and feed them into the spatial convolution of the UNet block. During inference, we use the average of the storyflows computed from the training set as the storyflow $\Delta$.

\setlength{\tabcolsep}{8pt}
\begin{table*}[t]
\scriptsize
\centering
\renewcommand\arraystretch{0.7}
    \begin{tabular}{lcccccc}
    \toprule 
    \multirow{2}{*}{\textbf{Model}} & \multicolumn{3}{c}{\textbf{PororoSV}} & \multicolumn{3}{c}{\textbf{FlintstonesSV}}\tabularnewline
    \cmidrule{2-7} 
     & \textbf{FID $\downarrow$}  & \textbf{Char. F1 $\uparrow$}  & \textbf{Frm. Acc. $\uparrow$}  & \textbf{FID $\downarrow$}  & \textbf{Char. F1 $\uparrow$}  & \textbf{Frm. Acc. $\uparrow$} \tabularnewline
    \midrule 
    StoryGANc  & 74.63  & 39.68  & 16.57  & 90.29  & 72.80  & 58.39 \tabularnewline
    StoryDALL-E  & 25.90  & 36.97  & 17.26  & 26.49  & 73.43  & 55.19 \tabularnewline
    MEGA-StoryDALL-E  & 23.48  & 39.91  & 18.01  & 23.58  & 74.26  & 54.68 \tabularnewline
    Story-LDM  & 26.64 & 47.56 & 29.19 & 24.24 & 76.59 & 57.19 \tabularnewline
    AR-LDM  & 17.40  & -  & -  & 19.28  & -  & - \tabularnewline
    Causal-Story  & 16.98  & - & - & 19.03  & - & - \tabularnewline
    StoryImager  & 15.45  & - & - & 18.32  & - & - \tabularnewline
    RCDMs  & 16.25 & 59.03 & 41.48 & 14.96 & 85.51 & 78.44 \tabularnewline
    \textbf{ContextualStory}  & \textbf{13.86}  & \textbf{76.25}  & \textbf{50.72}  & \textbf{13.27}  & \textbf{91.29}  & \textbf{81.91} \tabularnewline
    \bottomrule
    \end{tabular}
  \caption{Quantitative comparison with SOTA methods of story continuation on PororoSV and FlintstonesSV.}
  \label{tab:sc}
\end{table*}	

\setlength{\tabcolsep}{8pt}
\begin{table}[t]
\scriptsize
\centering
\renewcommand\arraystretch{0.7}
    \begin{tabular}{lccc}
    \toprule
    \textbf{Model}     & \textbf{FID $\downarrow$}            & \textbf{Char. F1 $\uparrow$}       & \textbf{Frm. Acc. $\uparrow$}      \\
    \midrule
    \textbf{PororoSV} \\
    \midrule
    StoryGAN & 158.06         & 18.59          & 9.34           \\
    CP-CSV & 149.29         & 21.78          & 10.03          \\
    DUCO & 96.51          & 38.01          & 13.97          \\
    VLC & 84.96          & 43.02          & 17.36          \\
    VP-CSV & 65.51          & 56.84          & 25.87          \\
    Word-Level SV & 56.08          & -              & -              \\
    Story-LDM & 27.33 & - & - \\
    AR-LDM & 16.59          & -              & -              \\
    Causal-Story & 16.28  & - & - \\
    StoryImager & 15.63  & - & - \\
    \textbf{ContextualStory} & \textbf{13.61} & \textbf{77.24} & \textbf{51.59} \\ 
        \midrule
        \textbf{FlintstonesSV} \\
        \midrule
        StoryGAN & 127.19         & 46.20              & 32.96             \\
        DUCO & 78.02          & 54.92              & 36.34             \\
        VLC & 72.87          & 58.81              & 39.18             \\
        Story-LDM & 36.55  & - & - \\
        AR-LDM & 23.59  & - & - \\
        StoryImager & 22.27  & - & - \\
        \textbf{ContextualStory} & \textbf{20.15} & \textbf{91.70}     & \textbf{83.08}   \\ 
    \bottomrule
    \end{tabular}
  \caption{Quantitative comparison with SOTA methods of story visualization on PororoSV and FlintstonesSV.}
  \label{tab:sv}
\end{table}

\subsection{Solving the Story Continuation Task}
\label{sec:Solving_the_Story_Continuation_Task}

For story continuation tasks, besides the storyline, the first frame is provided as an additional input. As shown in Figure \ref{fig:framwork_sc}, we extract the first frame latent $\mathbf{z^1}$ and use it as an additional input to all UNet blocks. Within each UNet block, $\mathbf{z^1}$ is resized to match the spatial dimensions of the hidden state, then a $1 \times 1$ convolution layer adjusts the channels to match the hidden state. Finally, it is concatenated with the hidden state before inputting into the spatial convolution.

\section{Experiment}

\subsection{Experimental Setup}
\noindent\textbf{Datasets.}
We employ two popular benchmark datasets, PororoSV \cite{li2019storygan} and  FlintstonesSV \cite{gupta2018imagine}, to evaluate the performance of our model on story visualization and story continuation tasks. PororoSV contains 10,191, 2,334, and 2,208 stories within the train, validation, and test splits, respectively, featuring 9 main characters. FlintstonesSV contains 20,132, 2,071, and 2,309 stories within the train, validation, and test splits, respectively, featuring 7 main characters and 323 backgrounds. Each story in  both datasets comprises 5 consecutive story images.

\noindent\textbf{Automatic metrics.}
To evaluate the quality of generated images, we employ the following three evaluation metrics following previous works \cite{maharana2022storydall,pan2024synthesizing} in story visualization: 
(1) Frechet Inception Distance (FID) \cite{heusel2017gans}, which measures the distance between feature vectors of ground truth and generated frames;
(2) Frame accuracy (Frm. Acc.), which evaluates character matching to ground truth using a fine-tuned Inception-v3 model;
(3) Character F1-score (Char. F1), which assesses the quality of generated characters using the same Inception-v3 model as Frm. Acc.

\noindent\textbf{Implementation details.}
We initialize ContextualStory with the pre-trained Stable Diffusion 2.1-base and fine-tune only the UNet parameters with the AdamW optimizer. Training is performed on 4 NVIDIA A800 GPUs with a batch size of 12, a learning rate of \(5 \times 10^{-5}\)
and 40,000 iterations for PororoSV and 80,000 iterations for FlintstonesSV. The SETA window size is \(k=3\), and the SC layer count is 4. 
During training, we apply classifier-free guidance by randomly dropping input storylines with a 0.1 probability and use the PYoCo mixed noise prior for noise initialization. For inference, we use the DDIM sampler with 50 steps and a guidance scale of 7.5 to generate $256 \times 256$ images.

\begin{figure*}[t]
    \centering
    \includegraphics[width=0.82\linewidth]{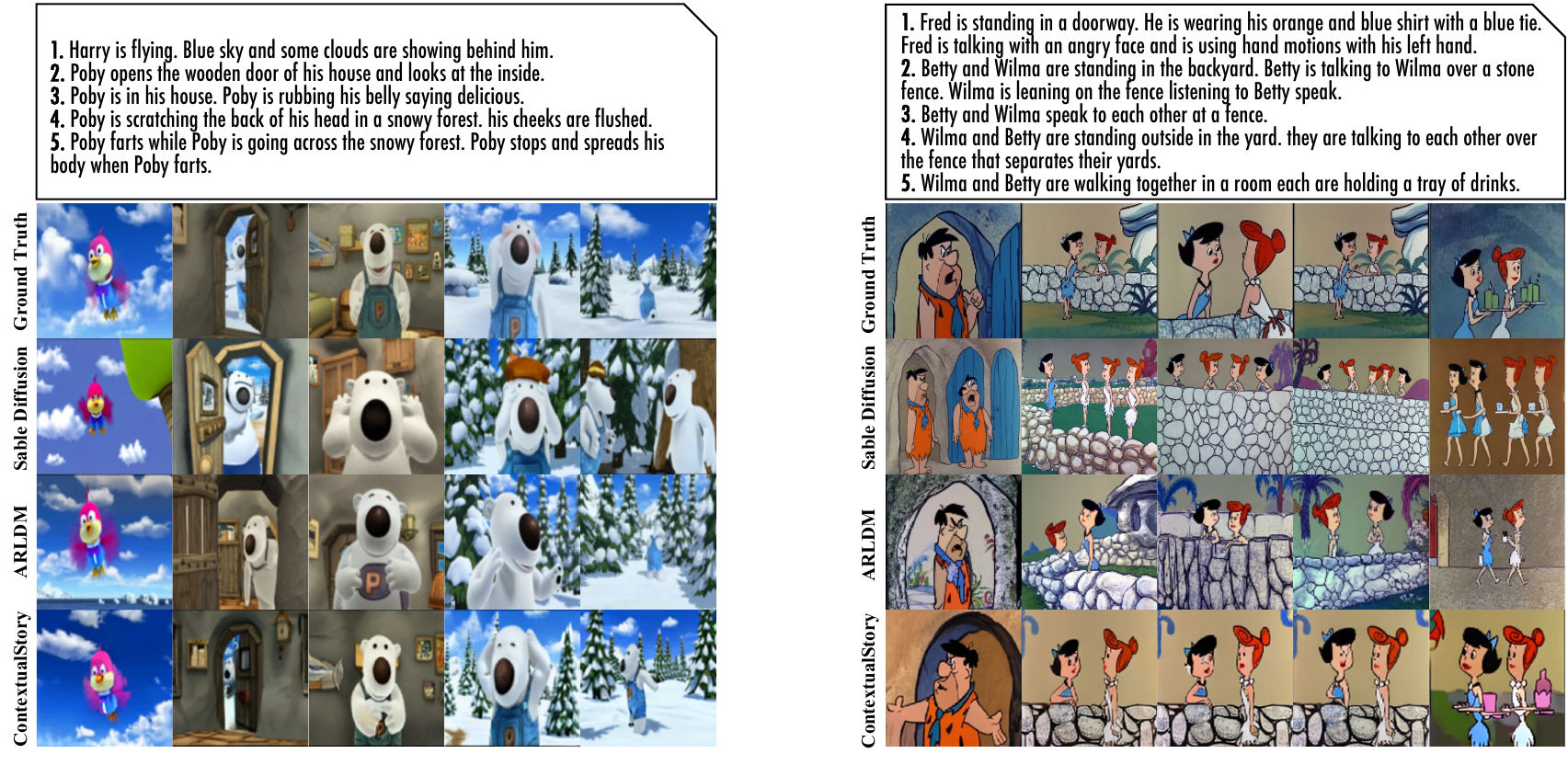}
    \caption{Qualitative comparison of story visualization on PororoSV (left) and FlintstonesSV (right).
    }
    \label{fig:result_sv}
\end{figure*}

\begin{figure*}[t]
    \centering
    \includegraphics[width=0.82\linewidth]{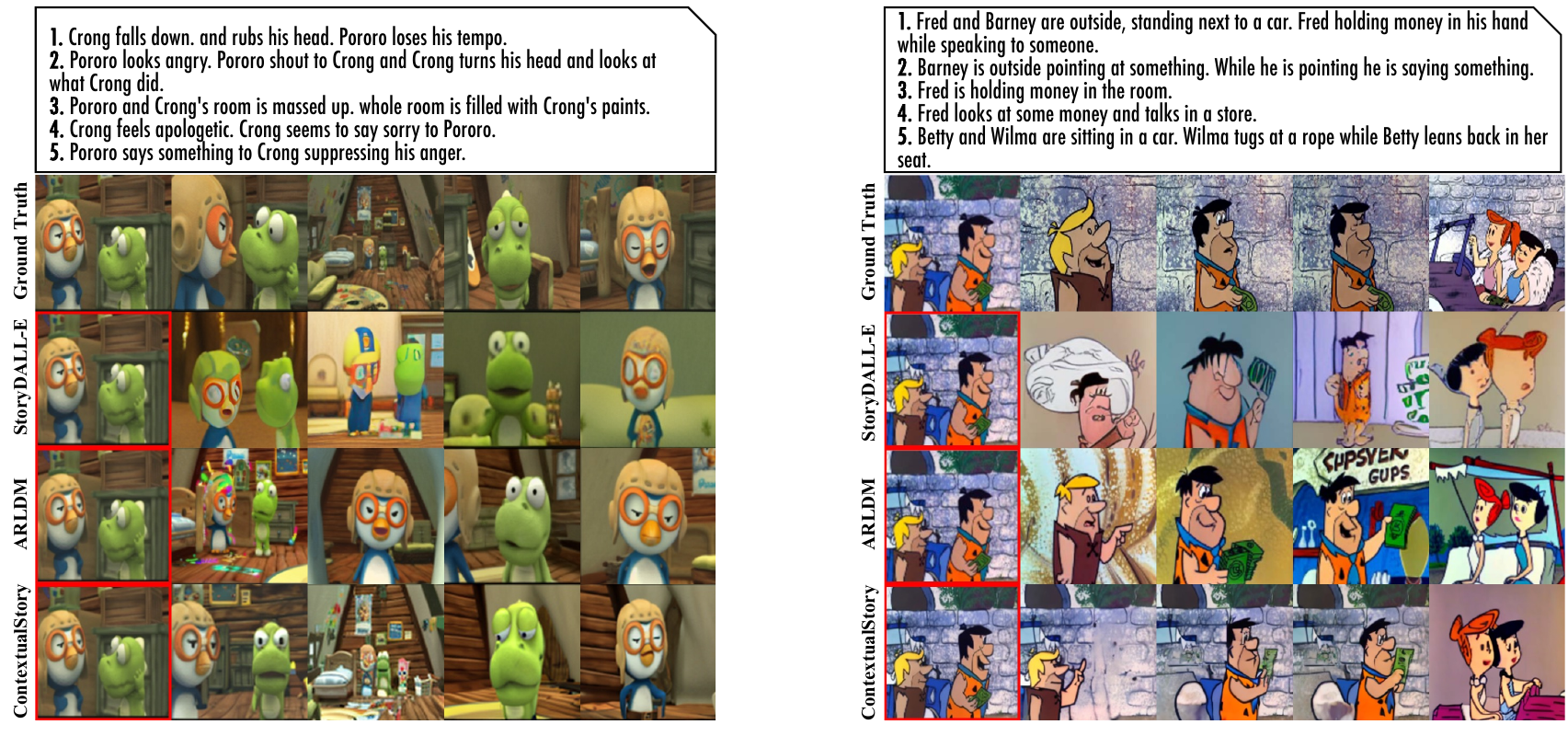}
    \caption{Qualitative comparison of story continuation on PororoSV (left) and FlintstonesSV (right). The image marked with a red box is the first frame additionally input to the model.
    }
    \label{fig:result_sc}
\end{figure*}

\setlength{\tabcolsep}{8pt}
\begin{table}[t]
\scriptsize
\centering
\renewcommand\arraystretch{0.7}
\begin{tabular}{lcc}
\toprule
\textbf{Model} & \textbf{Memory (GB)$\downarrow$} & \textbf{Inference Speed (s)$\downarrow$} \\ 
\midrule
StoryDALL-E    & 20                       & 347                              \\ 
Story-LDM      & 11                       & 18.5                             \\ 
AR-LDM          & 40                       & 40.4                             \\ 
StoryGen       & 29                       & 31.7                             \\ 
StoryGPT-V     & 25                       & 14.1                             \\ 
RCDMs          & 22                       & 30.4                             \\ 
\textbf{ContextualStory}  & \textbf{5}               & \textbf{11.8}                    \\ 
\bottomrule
\end{tabular}
\caption{Comparison of GPU memory usage and inference speed across SOTA models.}
\label{tab:memory_speed_comparison}
\end{table}	

\setlength{\tabcolsep}{5pt}
\begin{table}[t]
  \scriptsize
  \centering
  \renewcommand\arraystretch{0.7}
\begin{tabular}{lcccc}
\toprule 
\textbf{Dataset} & \textbf{Attribute} & \textbf{Ours} & \textbf{Tie} & \textbf{AR-LDM}\\
\midrule 
\multirow{3}{*}{PororoSV} & Visual Quality & \textbf{81.0\%} & 6.9\% & 12.1\%\\
 & Semantic Relevance & \textbf{85.6\%} & 9.2\% & 5.2\%\\
 & Temproal Consistency & \textbf{84.1\%} & 8.8\% & 7.1\%\\
\midrule 
\multirow{3}{*}{FlintstonesSV} & Visual Quality & \textbf{80.4\%} & 6.2\% & 13.4\%\\
 & Semantic Relevance & \textbf{82.6\%} & 6.3\% & 11.1\%\\
 & Temproal Consistency & \textbf{84.8\%} & 5.4\% & 9.8\%\\
\bottomrule
\end{tabular}
  \caption{
  Human evaluations of story visualization task. Ours (\%) means our ContextualStory is preferred over AR-LDM. AR-LDM (\%) means AR-LDM is preferred over our ContextualStory. Tie (\%) means the annotator believes that the two image sequences are similar.
  }
  \label{tab:human_evaluation}
\end{table}

\begin{figure}[t]
    \centering
    \includegraphics[width=0.81\linewidth]{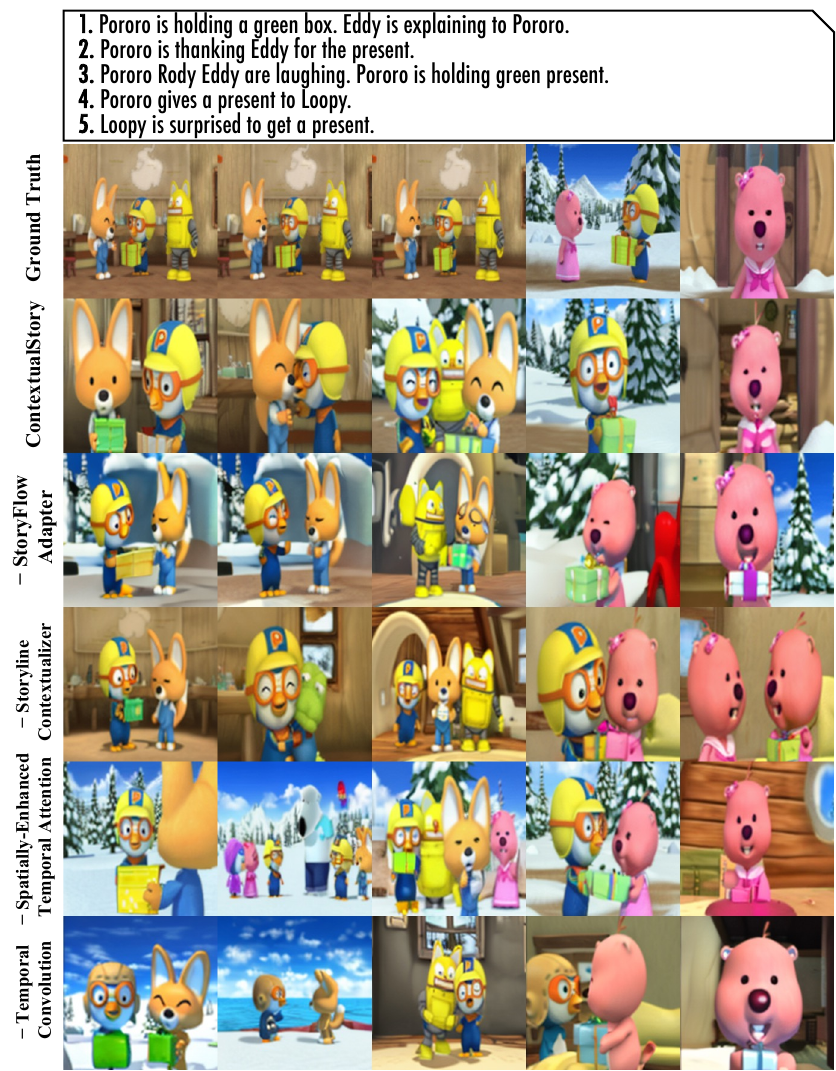}
    \caption{Qualitative results of the ablation study on the proposed components for story visualization on PororoSV.
    }
    \label{fig:ablation}
\end{figure}

\subsection{Quantitative Results}
\noindent\textbf{Story Visualization.}
Table \ref{tab:sv} presents quantitative results for story visualization on both PororoSV and FlintstonesSV, comparing ContextualStory to several SOTA methods, including StoryGAN, CP-CSV~\cite{song2020character}, DUCO~\cite{maharana2021improving}, VLC~\cite{maharana2021integrating}, VP-CSV~\cite{chen2022character}, Word-Level SV~\cite{li2022word}, Story-LDM, AR-LDM, Causal-Story~\cite{song2024causal}, and StoryImager. 
The results clearly demonstrate that ContextualStory significantly outperforms existing SOTA methods across all metrics on both datasets. This superior performance is primarily due to SETA, SC, and StoryFlow Adapter, which effectively utilize context information to generate coherent story frames.

\noindent\textbf{Story Continuation.}
Table \ref{tab:sc} presents the quantitative results for story continuation on both PororoSV and FlintstonesSV. 
We evaluate the effectiveness of ContextualStory model against several SOTA methods, such as StoryDALL-E, MEGA-StoryDALL-E~\cite{maharana2022storydall}, StoryImager, and RCDMs.
The results demonstrate that ContextualStory outperforms existing methods by a large margin across all metrics for the story continuation on both datasets. This indicates that ContextualStory better utilize contextual information to generate coherent story frames based on the storyline and the first image.

\noindent\textbf{Inference Speed.}
We compare the GPU memory usage and inference speed of recent open-source SOTA models (e.g., AR-LDM and StoryGen) in Table \ref{tab:memory_speed_comparison}. 
The experiment is conducted on an A800 GPU with 50 DDIM steps to ensure a fair comparison. 
Autoregressive methods like Story-LDM, AR-LDM, StoryGen, and StoryGPT-V suffer from high memory usage and slow inference speeds. 
In contrast, ContextualStory, a non-autoregressive model, not only overcomes the bottleneck of autoregressive methods by achieving the lowest memory usage and fastest inference speed, but also outperforms SOTA methods in overall performance.

\subsection{Qualitative Results}
\noindent\textbf{Story Visualization.}
Figure \ref{fig:result_sv} shows a qualitative comparison of story visualization on PororoSV and FlintstonesSV. Stable Diffusion (SD) generates high-quality images independently from individual sentences, but its lack of contextual awareness leads to inconsistent character appearances and character duplication. AR-LDM avoids character duplication but still struggles with inconsistent character appearances. In contrast, ContextualStory produces high-quality images with coherent and consistent characters and scenes across both datasets. 
% More results are provided in the supplementary material.

\noindent\textbf{Story Continuation.}
Figure \ref{fig:result_sc} demonstrates a qualitative comparison of story continuation on PororoSV and FlintstonesSV datasets. StoryDALL-E produces low-quality characters with inconsistent backgrounds. AR-LDM generates higher-quality characters, but the backgrounds lack consistency and deviate significantly from the ground truth. In contrast, ContextualStory generates high-quality images with consistent characters and backgrounds that closely match the ground truth. More results are provided in the supplementary material.

\subsection{Human Evaluation}
Due to the limitations of metrics such as FID, Char. F1, and Frm. Acc. in accurately reflecting the quality of generated story frames, we conducted human evaluations for the story visualization task on PororoSV and FlintstonesSV, focusing on \textit{Visual Quality}, \textit{Semantic Relevance}, and \textit{Temporal Consistency}.
We randomly selected 300 pairs of story frame sequences generated from AR-LDM  \cite{pan2024synthesizing} and our ContextualStory. Annotators were tasked to select the better sequence for the three attributes: Visual Quality, Semantic Relevance, and Temporal Consistency. Each pair of story frame sequences was evaluated by 10 annotators. 
As shown in Table \ref{tab:human_evaluation}, the results indicate that our ContextualStory outperforms AR-LDM significantly across all three attributes.

\subsection{Ablation Studies}
\noindent\textbf{Ablation study of the proposed components.}
To evaluate the benefit of each proposed component, we conduct an ablation study on the story visualization task using PororoSV. 
As shown in Table \ref{tab:ablation}, progressively removing components from ContextualStory results in a consistent decline across all three metrics. The removal of SETA has the most significant effect, increasing FID by 16.0\%, and reducing Char. F1 and Frm. Acc. by 4.9\% and 7.4\%, respectively.
The qualitative comparison in Figure \ref{fig:ablation} shows the following:
1) Removing the StoryFlow Adapter slightly reduces background consistency.
2) Further removing SC leads to duplicated characters, like Loopy.
3) Removing SETA reduces background consistency, introduces duplicated characters (\textit{e.g.}, Pororo), and incorrect characters (\textit{e.g.}, Petty and Poby), making images less accurate.
4) Removing Temporal Convolution further decreases character and scene consistency.
These results indicate that all proposed components contribute to the performance of ContextualStory, with SETA having the most significant impact.

\noindent\textbf{Ablation study of temporal attention.}
Table \ref{tab:ablation} presents the ablation study results of comparing Vanilla Temporal Attention and our proposed SETA for the story visualization task on PororoSV. The results clearly show that SETA outperforms Vanilla Temporal Attention across all metrics. Specifically, SETA achieves a lower FID score, indicating better alignment with ground truth images, and higher Char. F1 and Frm. Acc., demonstrating improved character consistency and accuracy. These improvements highlight the effectiveness of the local window mechanism of SETA, which allows the model to better capture both spatial and temporal dependencies, leading to more coherent and consistent story frames. The significant gains in performance suggest that incorporating spatial context within temporal attention is crucial for enhancing visual storytelling models. More ablation studies are provided in the supplementary material.

\setlength{\tabcolsep}{8pt}
\begin{table}[t]
\scriptsize
\centering
\renewcommand\arraystretch{0.7}
        \begin{tabular}{lccc}
            \toprule
            \textbf{Model}    & \textbf{FID $\downarrow$}            & \textbf{Char. F1 $\uparrow$}       & \textbf{Frm. Acc. $\uparrow$} \\
            \midrule
            \textbf{ContextualStory}     & \textbf{13.61} & \textbf{77.24}    & \textbf{51.59} \\
            \ $-$StoryFlow Adapter & 14.84 & 77.09 & 50.48 \\
            \ \ $-$Storyline Contextualizer & 15.02 & 75.42 & 48.39 \\
            \ \ \ $-$SETA & 17.42 & 71.70 & 44.83 \\  
            \ \ \ \ $-$Temporal Convolution & 19.69 & 68.12 & 39.60 \\ 
            \midrule
            Vanilla Temporal Attention & 14.78 & 75.94 & 48.79\\
            \textbf{SETA (Ours)} & \textbf{13.61} & \textbf{77.24} & \textbf{51.59}\\
            \bottomrule
        \end{tabular}
      \caption{Ablation study of the proposed components and temporal attention for story visualization on PororoSV.}
      \label{tab:ablation}
\end{table}

\section{Conclusion}
In this paper, we propose ContextualStory, a novel framework that overcomes the limitations of traditional autoregressive methods in visual storytelling. By incorporating Spatially-Enhanced Temporal Attention, we effectively capture spatial and temporal dependencies, ensuring consistency in characters and scenes across frames. Additionally, the Storyline Contextualizer enriches the global context from storyline, while the StoryFlow Adapter enhances the ability of model to handle scene changes. Extensive experiments on PororoSV and FlintstonesSV show that ContextualStory achieves superior performance over state-of-the-art methods in both story visualization and continuation tasks.

\appendix
\section{Acknowledgments}
This paper is supported by the Doubao Fund.

{
    % \small
    \bibliography{aaai25}
}

\newpage

\part*{Supplementary Material}

This supplementary material provides additional details to complement the main paper. It includes more related works, a detailed explanation of the UNet architecture and training process, and descriptions of the PororoSV and FlintstonesSV datasets. We also present more ablation studies, comparing different model components and configurations. Additionally, we provide more quantitative and qualitative results, demonstrating the effectiveness of ContextualStory. Finally, we discuss the limitations of our approach and suggest potential improvements for future work.

\section{More Related Works}
\noindent \textbf{Text-to-video generation.}
Text-to-video generation focuses on creating temporally coherent videos \cite{brooks2022generating,ge2022long,saito2020train,yu2022generating,le2021ccvs,wu2021godiva}. The success of diffusion models in text-to-image generation has significantly advanced this field \cite{ho2022video,singer2022make,ho2022imagen,blattmann2023stable,an2023latent,zhou2022magicvideo,he2022latent,mei2023vidm,yu2023video,bain2021frozen,blattmann2023align,guo2023animatediff,luo2023videofusion,wang2023videofactory,yin2023nuwa}. Pioneering approaches like VDM \cite{ho2022video} use a space-time factorized UNet with joint image and video  training. Make-a-Video \cite{singer2022make} and Imagen Video  \cite{ho2022imagen} capture video distribution at low resolution before enhancing resolution and duration through spatiotemporal interpolation.
Leveraging the efficiency of LDM, subsequent works \cite{zhou2022magicvideo,he2022latent,blattmann2023align,mei2023vidm,yu2023video,bain2021frozen,wang2023lavie,wang2023modelscope,guo2023animatediff} extended 2D UNet by interleaving temporal layers between pre-trained 2D layers and fine-tuning on large-scale video datasets. LaVie  \cite{wang2023lavie} and ModelScopeT2V  \cite{wang2023modelscope} fine-tune the entire model, while VideoLDM \cite{blattmann2023align} and AnimateDiff  \cite{guo2023animatediff} fine-tune only additional temporal layers, making them plug-and-play for personalized image models.
Diffusion transformers (DiT) \cite{peebles2023scalable,bao2023all,ma2024sit} have revolutionized video generation, leading to sophisticated solutions like Latte \cite{ma2024latte}, W.A.L.T.  \cite{gupta2023photorealistic}, and Sora \cite{videoworldsimulators2024}. These methods extract spacetime patches from input videos and use DiTs to model video distribution in latent space. While Sora excels at generating minute-long videos, the content often covers a limited range of scenes or simple motion changes. In contrast, story visualization focuses on generating images corresponding to multiple sentences, ensuring global consistency of dynamic scenes and characters.

\begin{figure}[t]
    \centering
    \includegraphics[width=\linewidth]{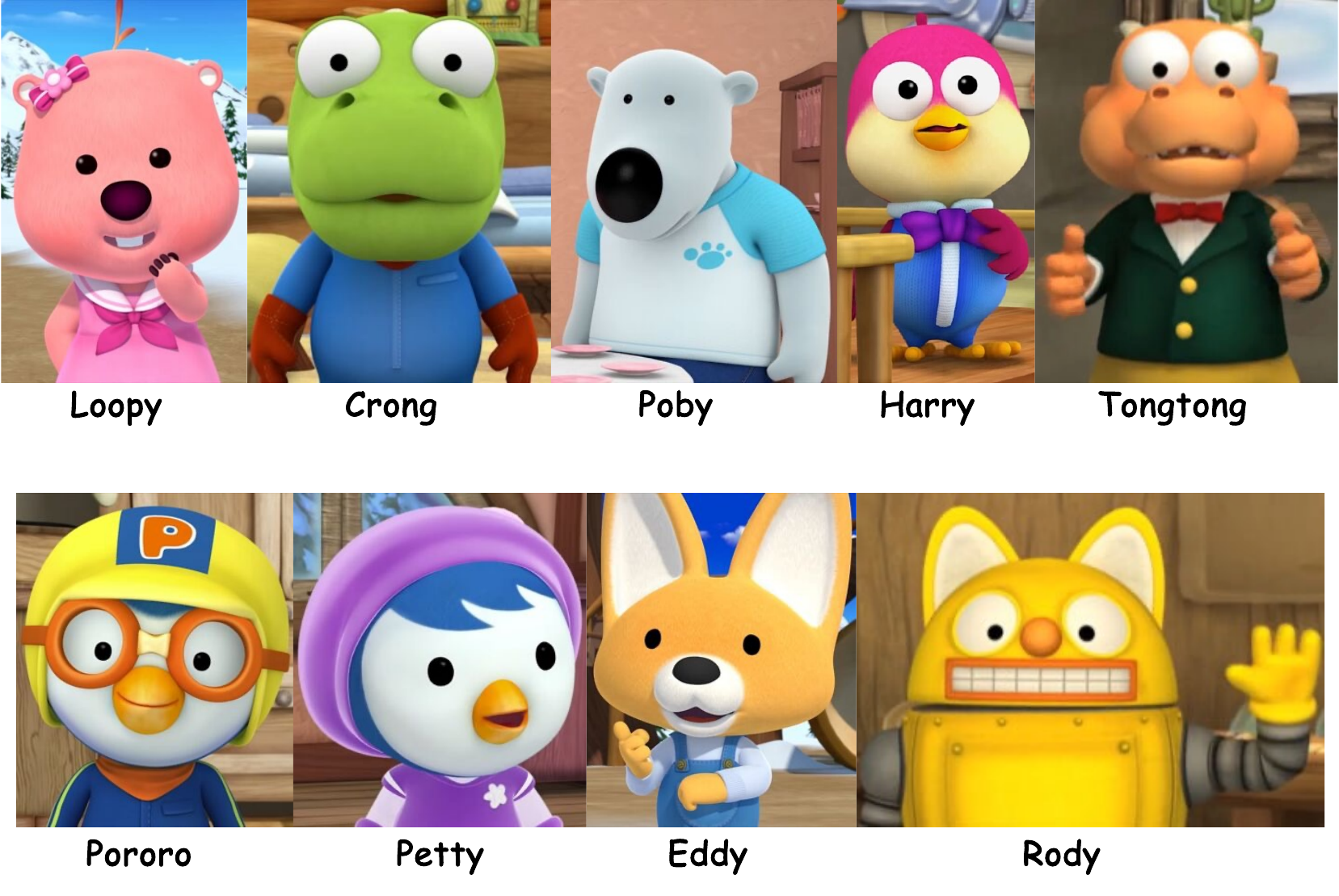}
    \caption{Main character names and their corresponding images in PororoSV. The images are sourced from https://pororo.fandom.com/.}
    \label{fig:main_character_pororo}
\end{figure}

\begin{figure}[t]
    \centering
    \includegraphics[width=\linewidth]{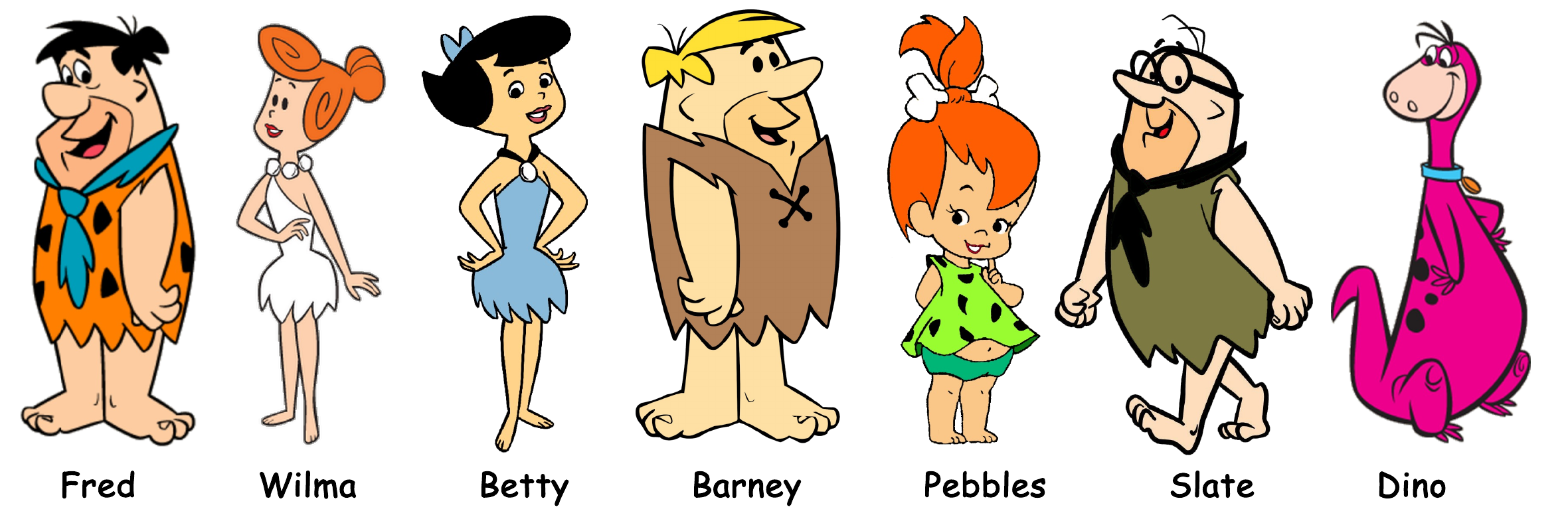}
    \caption{Main character names and their corresponding images in FlintstonesSV. The images are sourced from https://flintstones.fandom.com/.}
    \label{fig:main_character_flintstones}
    \vspace{-5pt}
\end{figure}

\section{UNet Architecture}

Our ContextualStory is developed based on the text-to-image generation model Stable Diffusion 2.1-base, which utilizes UNet for the diffusion and denoising processes in the latent space to generate images. The UNet consists of four downsampling blocks, one middle block, and four upsampling blocks. Each block contains a spatial convolution block. Apart from the last downsampling block and the first upsampling block, each block also includes a spatial attention block.

\setlength{\tabcolsep}{8pt}
\begin{table*}
    \scriptsize
    \centering
    % \vspace{-5pt}
\begin{tabular}{lccccccc}
\toprule
\multirow{2}{*}{\textbf{Model}} & \multirow{2}{*}{\textbf{w/ ref. text}} & \multicolumn{3}{c}{\textbf{PororoSV}}                 & \multicolumn{3}{c}{\textbf{FlintstonesSV}}            \\ \cmidrule{3-8}
                                &                                        & \textbf{FID $\downarrow$} & \textbf{Char. F1 $\uparrow$} & \textbf{Frm. Acc. $\uparrow$} & \textbf{FID $\downarrow$} & \textbf{Char. F1 $\uparrow$} & \textbf{Frm. Acc. $\uparrow$} \\ \midrule
\multicolumn{8}{c}{\textbf{Story Visualization}}                                                                                                                                         \\ \midrule
ContextualStory                   & \ding{51}                                      & 14.28        & 69.65             & 39.62              & 21.94 & 88.67 & 76.64             \\
ContextualStory                   & \ding{55}                                      & 13.61        & 77.24             & 51.59              & 20.15    & 91.70    & 83.08              \\ \midrule
\multicolumn{8}{c}{\textbf{Story Continuation}}                                                                                                                                          \\ \midrule
ContextualStory                   & \ding{51}                                      & 15.47        & 69.82             & 39.86              & 14.87 & 88.85 & 78.22              \\
ContextualStory                   & \ding{55}                                      & 13.86        & 76.25             & 50.72              & 13.27        & 91.29             & 81.91              \\ \bottomrule
\end{tabular}
    \captionof{table}{Results of our ContextualStory for story visualization and story continuation tasks with and without the reference text on the PororoSV and FlintstonesSV datasets.}
    \label{tab:with_reference_ablation_sv_sc}
    % \vspace{-5pt}
\end{table*}

\setlength{\tabcolsep}{8pt}
\begin{table*}
    \scriptsize
    \centering
    % \vspace{-5pt}
    \begin{tabular}{lccccccc}
\toprule
\multirow{2}{*}{\textbf{Model}} & \multirow{2}{*}{\textbf{w/ ref. text}} & \multicolumn{3}{c}{\textbf{PororoSV}}                   & \multicolumn{3}{c}{\textbf{FlintstonesSV}}            \\  \cmidrule{3-8}
                                &                                        & \textbf{FID $\downarrow$}   & \textbf{Char. F1 $\uparrow$} & \textbf{Frm. Acc. $\uparrow$} & \textbf{FID $\downarrow$} & \textbf{Char. F1 $\uparrow$} & \textbf{Frm. Acc. $\uparrow$} \\  \midrule
StoryDALL-E                     & \multirow{5}{*}{\ding{51}}                     & 40.39          & 50.56             & 21.03              & 44.66        & 78.36             & 61.83              \\
LDM                             &                                        & 60.23          & 56.30             & 16.59              & 87.39        & 78.68             & 57.38              \\
Story-LDM                       &                                        & 36.64          & 57.95             & 20.26              & 69.49        & 86.59             & 69.19              \\
StoryGPT-V                      &                                        & 19.56          & 62.70             & 36.06              & \textbf{21.71}        & \textbf{94.17}             & \textbf{87.96}              \\
\textbf{ContextualStory (Ours)}   &                                        & \textbf{14.28} & \textbf{69.65}    & \textbf{39.62}     & 21.94        & 88.67             & 76.64     \\  \bottomrule        
\end{tabular}
    \captionof{table}{Quantitative comparison with the state-of-the-art methods for the story visualization task with reference text on the PororoSV and FlintstonesSV datasets.}
    \label{tab:with_reference_sv}
    \vspace{-5pt}
\end{table*}

While the original UNet captures spatial dependencies, temporal dependencies is crucial for enhancing the consistency in visual storytelling. Hence, we introduced temporal convolution and Spatially-Enhanced Temporal Attention (SETA) to UNet to effectively capture spatial and temporal dependencies for addressing inconsistency problem. 
Specifically, we add a temporal convolution block after every spatial convolution block and a SETA block after every spatial attention block.
Temporal convolution blocks have the same architecture as their corresponding spatial counterparts, with the key difference being that temporal convolution blocks operate along the temporal dimension.
The spatial attention block comprises a self-attention layer that operates independently on each story frame, and a cross-attention layer that operates between the story frames and the storyline embedding. 
The SETA blocks do not require the guidance of storyline embedding, so they do not contain a cross-attention layer but two self-attention layers.
Spatial convolution and temporal convolution capture spatiotemporal dependencies among the story frames by convolving over their spatial and temporal dimensions, while spatial attention and SETA capture spatiotemporal dependencies by selectively attending to different regions within the images and other images. Through the integration of these spatiotemporal blocks, our ContextualStory effectively captures the complex spatial and temporal dependencies within the story frames, resulting in the generation of coherent story frames.

For story continuation tasks, in addition to the storyline embedding, the first frame serves as an additional input. We modified the architecture of ContextualStory slightly to accommodate this. Specifically, we extract the latent representation of the first frame and utilize it as an additional guiding input to all UNet blocks. Within each UNet block, we first resize it to align with the spatial dimensions of the hidden state. Subsequently, we apply a $1 \times 1$ convolution layer to adjust the channel to match that of the hidden state before concatenating it with the hidden state. Finally, the concatenated feature is inputted into the spatial convolution.

\section{Details of Training}

\subsection{Training Objective}

\noindent\textbf{Story visualization.}
During training, the UNet inputs noise latent inputs and predicts the added noise under the guidance of storyline embedding, timestep embedding, and storyflow embedding. The training objective of ContextualStory is defined as:
% \vspace{-1pt}
\begin{equation}
L_{LDM}^{SV}:=\mathbb{E}_{\mathcal{E}(\mathcal{I}),\mathcal{C},\boldsymbol{\epsilon},t}[\|\epsilon-\epsilon_\theta(\mathcal{Z}_t,t,\mathcal{C}, \Delta ')\|^2].
\end{equation}

% \vspace{-3pt}
\noindent\textbf{Story continuation.}
In comparison to the story visualization task, the story continuation task provides the latent representation of the first frame as an additional input to the UNet during training.
Therefore, the training objective of ContextualStory is defined as:
\small{
\begin{equation}
L_{LDM}^{SC}:=\mathbb{E}_{\mathcal{E}(I^{2:N}),\mathbf{c}^{2:N},\boldsymbol{\epsilon},t}[\|\epsilon-\epsilon_\theta(\mathcal{Z}_t,t,\mathbf{c}^{2:N}, \Delta ')\|^2].
\end{equation}
}

\begin{figure*}[t]
    \centering
    \includegraphics[width=\linewidth]{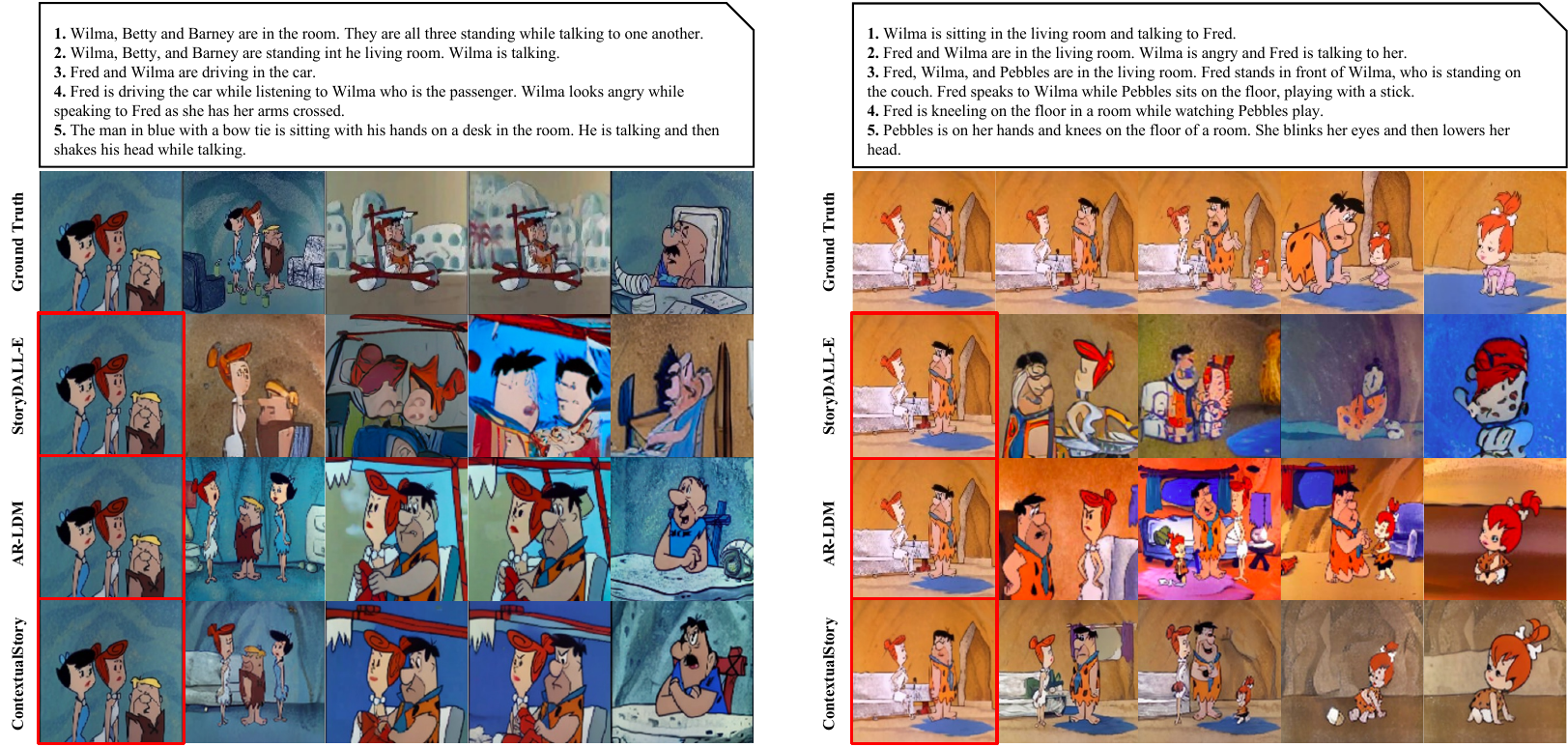}
    % \vspace{-5pt}
    \caption{Qualitative comparison of story continuation on PororoSV (left) and  FlintstonesSV (right). The image marked with a red box is the first frame additionally input to the model.}
    \label{fig:result_sc}
    \vspace{-5pt}
\end{figure*}

\begin{figure}[t]
    \centering
    \includegraphics[width=0.85\linewidth]{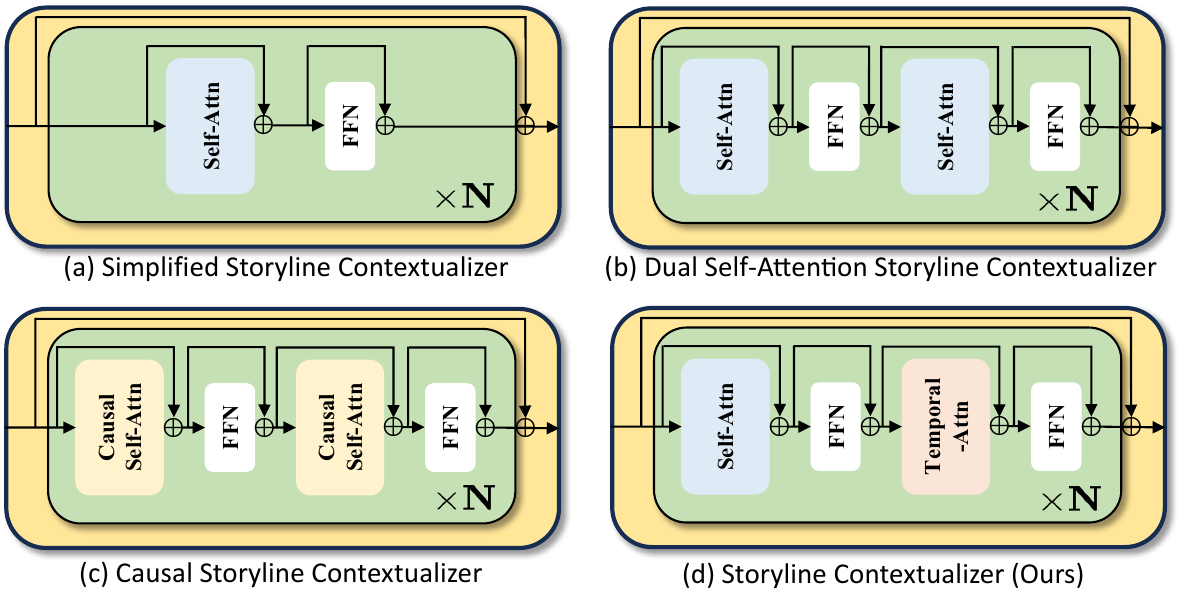}
    \caption{\textbf{Variants of the Storyline Contextualizer.} (a) \textbf{Simplified Storyline Contextualizer} consists of a single self-attention layer and a feed-forward network (FFN) repeated across N layers. (b) \textbf{Dual Self-Attention Storyline Contextualizer} adds an additional self-attention layer per block. (c) \textbf{Causal Storyline Contextualizer} replaces the self-attention layers with causal self-attention layers. (d) \textbf{Storyline Contextualizer (Ours)} incorporates temporal attention to capture temporal dependencies, alongside self-attention and FFN layers.}
    \label{fig:SC_variants}
    \vspace{-10pt}
\end{figure}

\section{Details of Datasets}

Following previous works~\cite{pan2024synthesizing,rahman2023make,shen2023large,wang2024evolving,shen2024boosting,tao2024storyimager}, we employ PororoSV \cite{li2019storygan} and the FlintstonesSV \cite{gupta2018imagine}, to evaluate the performance of our ContextualStory in both story visualization and story continuation tasks.

\noindent\textbf{PororoSV}
The PororoSV dataset comprises 10,191 training samples, 2,334 validation samples, and 2,208 test samples. Each sample is composed of a sequence of 5 frames paired with corresponding 5 sentences, forming a storyline. The dataset features 9 main characters: Loopy, Crong, Poby, Harry, Tongtong, Pororo, Petty, Eddy, and Rody. Figure \ref{fig:main_character_pororo} illustrates the profile images of these characters. This dataset is designed for the story visualization task, ensuring a diverse range of scenes and interactions between the characters.

\noindent\textbf{FlintstonesSV}
The FlintstonesSV dataset contains 20,132 samples for training, 2,071 for validation, and 2,309 for testing. Similar to PororoSV, each sample consists of a sequence of 5 frames paired with corresponding 5 sentences, forming a storyline. The 7 main characters featured in this dataset include Fred, Wilma, Betty, Barney, Pebbles, Slate, and Dino. Figure \ref{fig:main_character_flintstones} shows their profile images. The dataset was originally used for text-to-video synthesis and has been adapted for story visualization tasks to maintain consistency with prior research.

\section{More Quantitative Results}

\noindent\textbf{Results on extended dataset with reference text.}
Story-LDM~\cite{rahman2023make} pioneered the introduction of reference resolution in the story visualization task, proposing a framework based on autoregressive diffusion with a memory-attention module to address ambiguous references. 
Story-LDM extends the dataset by replacing character names with references, \textit{i.e.}, he, she, or they.
We trained our ContextualStory on the PororoSV and FlintstonesSV datasets with reference text for story visualization and story continuation tasks. As shown in Table \ref{tab:with_reference_ablation_sv_sc}, the performance of our ContextualStory on datasets with reference text only exhibits a slight decrease, indicating that even on more challenging datasets with reference text, our model can still effectively generate coherent story frames.
Furthermore, we compared our model with state-of-the-art methods, including StoryDALL-E~\cite{maharana2022storydall},  LDM~\cite{rombach2022high}, Story-LDM, and StoryGPT-V~\cite{shen2023large}. 
The results presented in Table \ref{tab:with_reference_sv} demonstrate that our model surpasses the existing state-of-the-art methods on the PororoSV dataset. Moreover, our performance on the FlintstonesSV dataset is on par with the current state-of-the-art methods. 
The performance of our model is slightly lower than that of StoryGPT-V. This could be attributed to StoryGPT-V leveraging the powerful reasoning capabilities of LLM for reference resolution, whereas we only utilize the CLIP text encoder to resolve ambiguous references. 
These results demonstrate the effectiveness of our ContextualStory in resolving ambiguous references. 

\setlength{\tabcolsep}{3pt}
\begin{table}[t]
\scriptsize
\centering
\renewcommand\arraystretch{1}
    \begin{tabular}{lccc}
        \toprule
        \textbf{Method} & \textbf{FID $\downarrow$}            & \textbf{Char. F1 $\uparrow$}       & \textbf{Frm. Acc. $\uparrow$} \\
        \midrule
        Simplified Storyline Contextualizer                  & 13.94 & 74.60    & 47.18     \\
        Dual Self-Attention Storyline Contextualizer               & 13.75          & 75.31             & 47.84             \\  
        Causal Storyline Contextualizer                  & 14.59 & 75.02    & 48.00     \\
        \textbf{Storyline Contextualizer (Ours)}              & \textbf{13.61}          & \textbf{77.24}             & \textbf{51.59}             \\  
        \bottomrule
    \end{tabular}
  \caption{Ablation study of Storyline Contextualizer for the story visualization task on the PororoSV.}
  \label{tab:ablation_sc}
  \vspace{-10pt}
\end{table}

\section{More Qualitative Results}

Figure \ref{fig:result_sc} presents a qualitative comparison of story continuation on the PororoSV and FlintstonesSV datasets. While StoryDALL-E generates characters of low quality with mismatched backgrounds, AR-LDM improves on character quality but still struggles with inconsistent backgrounds that significantly differ from the ground truth. In contrast, ours ContextualStory produces high-quality images with characters and backgrounds that are not only consistent but also closely align with the ground truth. 

\begin{figure*}
    \centering
    \includegraphics[height=\textheight,width=\linewidth]{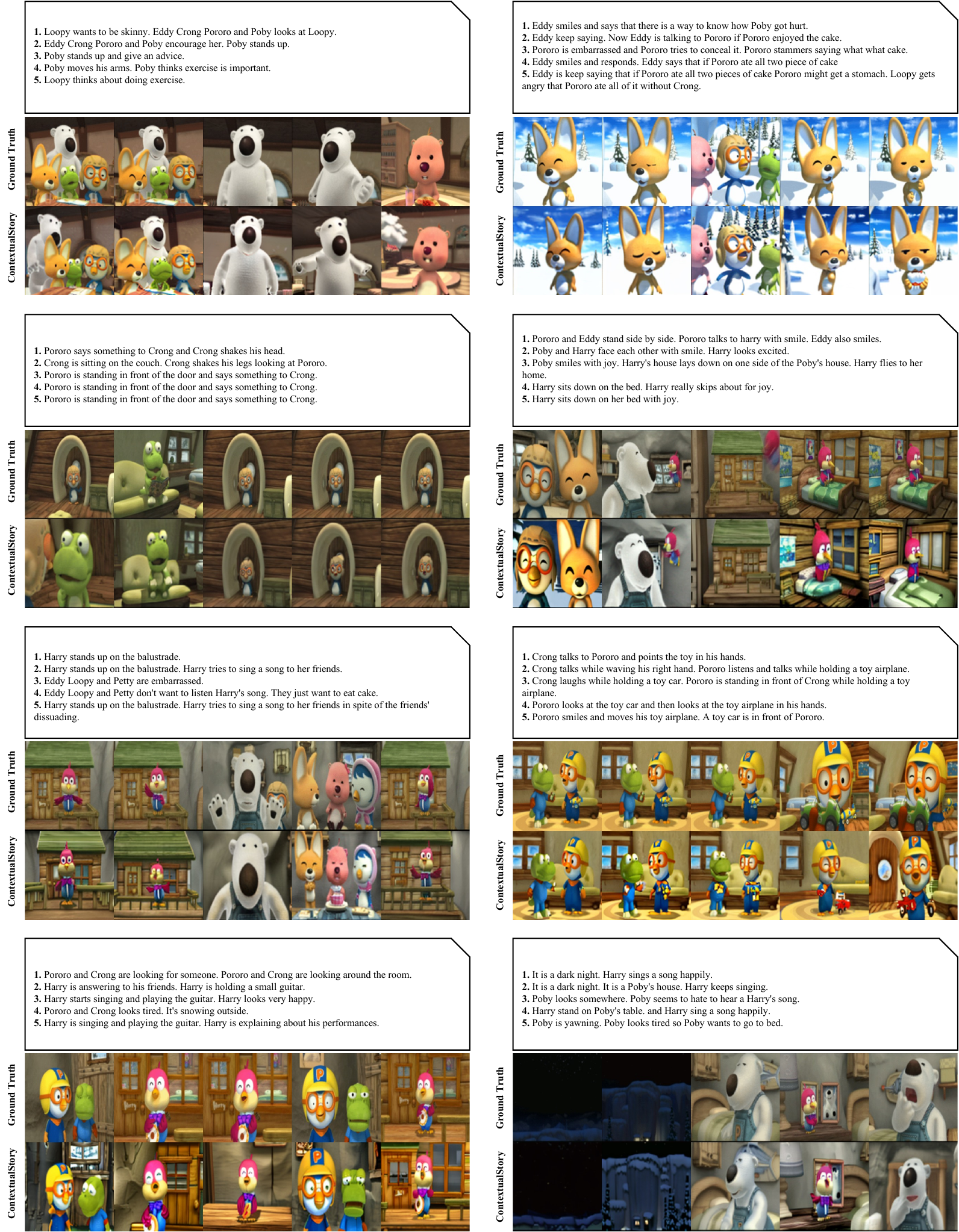}
    \caption{Qualitative results of the story visualization task on the PororoSV dataset.}
    \label{fig:supp_pororo_sv}
\end{figure*}

\begin{figure*}
    \centering
    \includegraphics[height=\textheight,width=\linewidth]{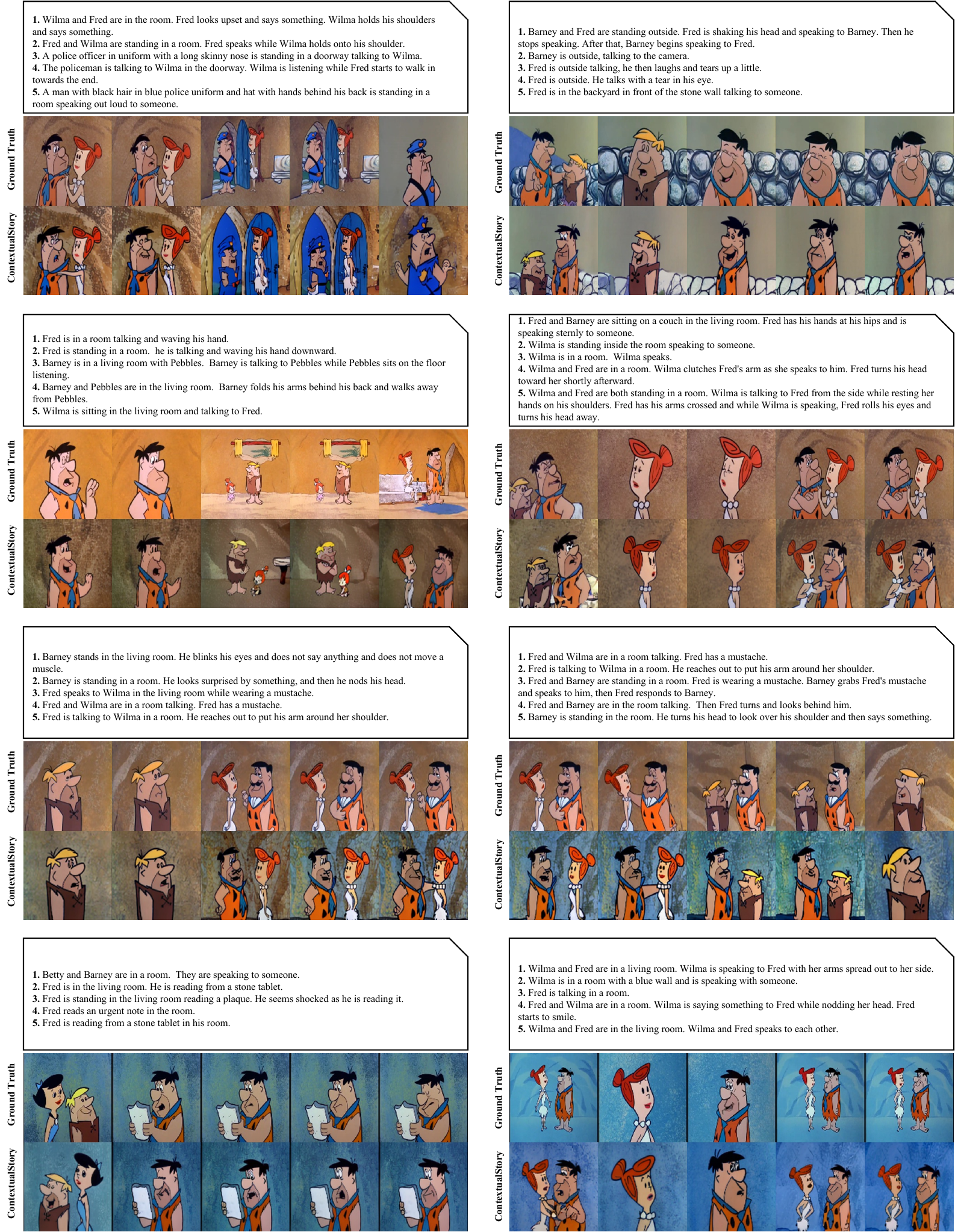}
    \caption{Qualitative results of the story visualization task on the FlintstonesSV dataset.}
    \label{fig:supp_flintstone_sv}
\end{figure*}

\begin{figure*}
    \centering
    \includegraphics[height=\textheight,width=\linewidth]{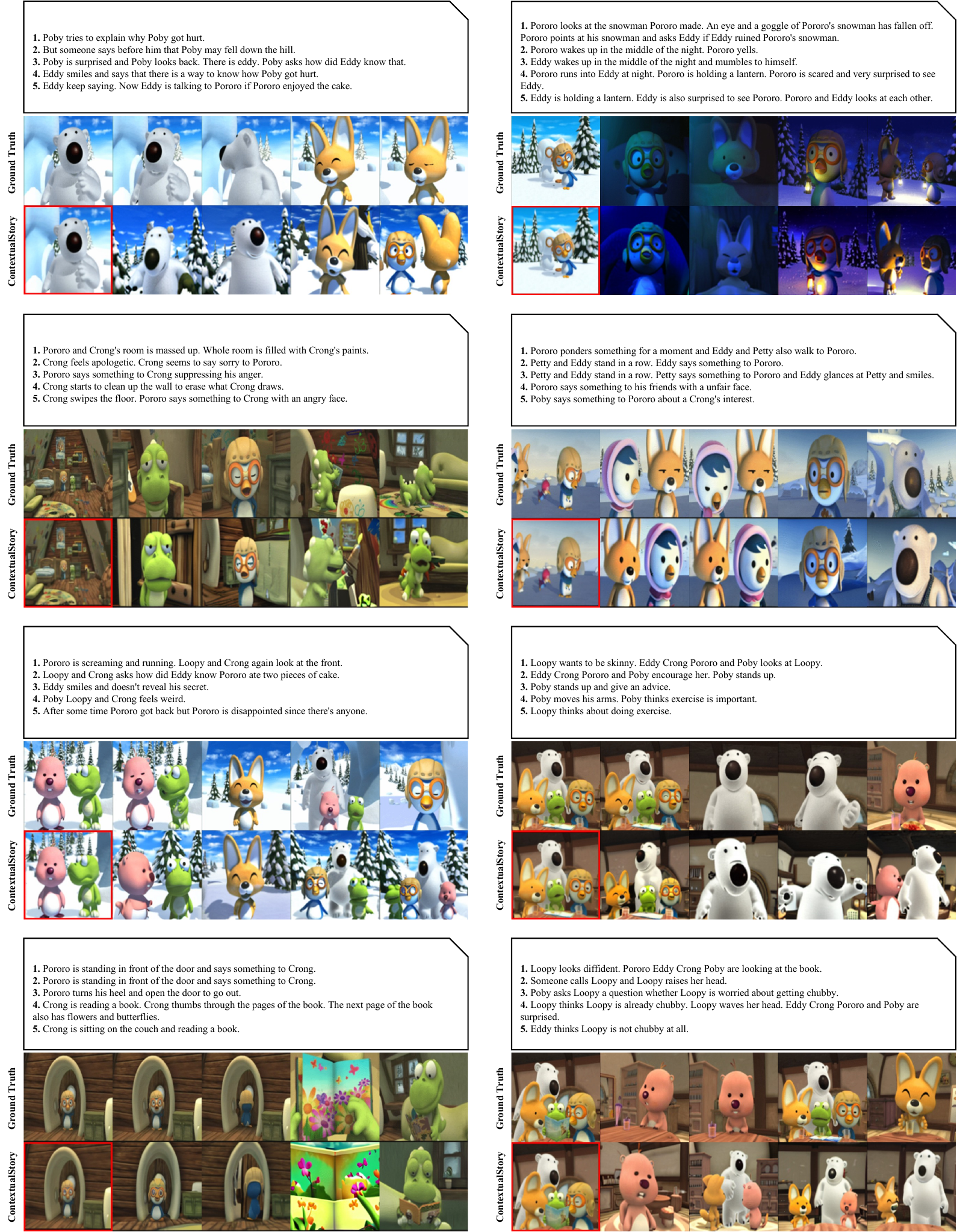}
    \caption{Qualitative results of the story continuation task on the PororoSV dataset.}
    \label{fig:supp_pororo_sc}
\end{figure*}

\begin{figure*}
    \centering
    \includegraphics[height=\textheight,width=\linewidth]{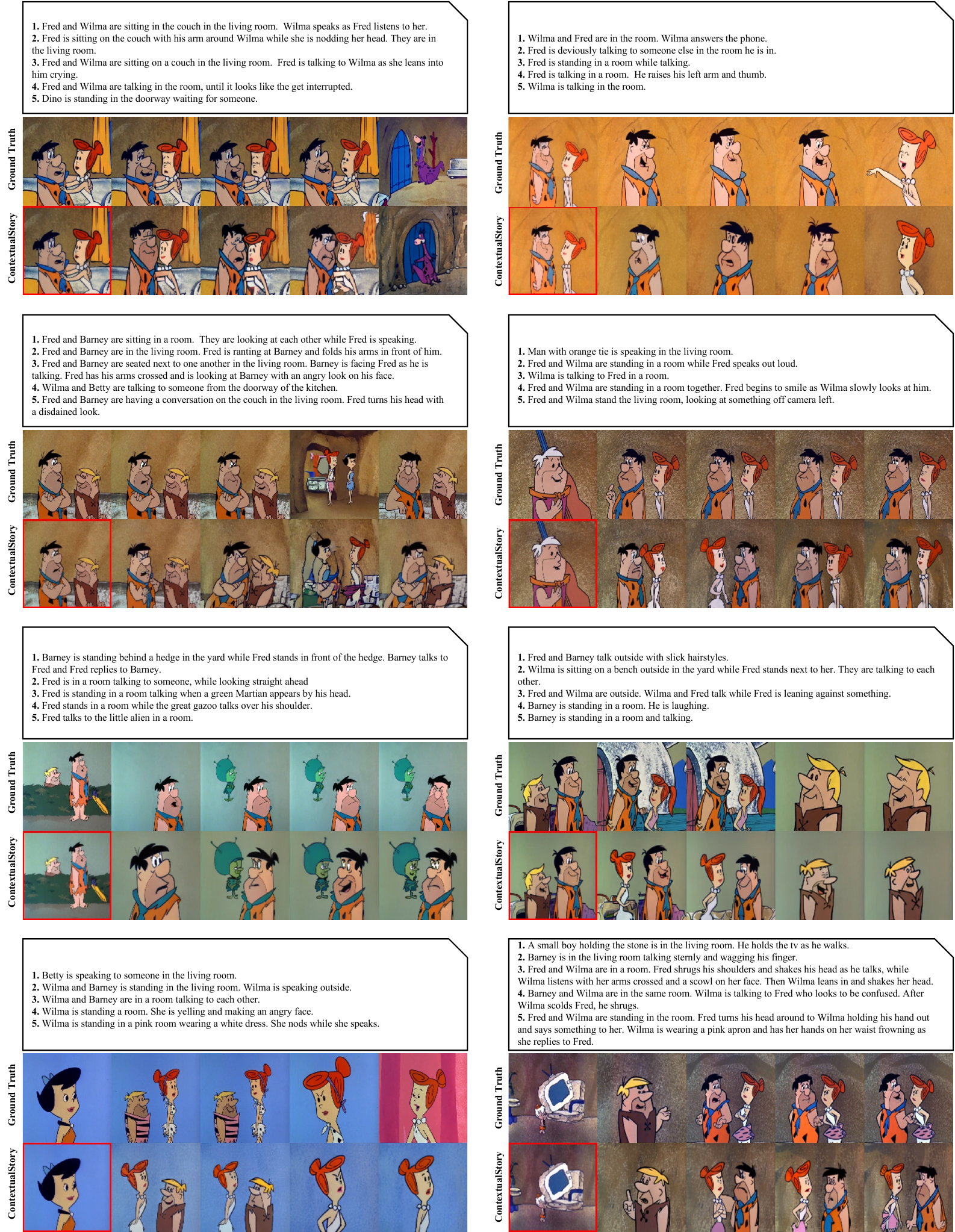}
    \caption{Qualitative results of the story continuation task on the FlintstonesSV dataset.}
    \label{fig:supp_flintstone_sc}
\end{figure*}

We also provide more qualitative results of the story visualization task on PororoSV and FlintstonesSV as shown in Figure \ref{fig:supp_pororo_sv} and Figure \ref{fig:supp_flintstone_sv}, and more qualitative results of the story continuation task on PororoSV and FlintstonesSV as shown in Figure \ref{fig:supp_pororo_sc} and Figure \ref{fig:supp_flintstone_sc}.
These results demonstrate the ability of ContextualStory to maintain character and scene consistency across story visualization and continuation tasks.

\section{More Ablation Studies}
\noindent\textbf{Ablation study of Storyline Contextualizer.}
The Storyline Contextualizer is evaluated in four different variants to assess their impact on visual storytelling tasks, as shown in Figure \ref{fig:SC_variants}. The Simplified Storyline Contextualizer employs a basic structure with a single self-attention layer followed by a feed-forward network (FFN). The Dual Self-Attention Storyline Contextualizer extends this by adding an additional self-attention layer. The Causal Storyline Contextualizer replaces the self-attention layers with causal self-attention layers. Finally, the proposed Storyline Contextualizer integrates temporal attention, enhancing the model's ability to capture temporal dependencies.

We conduct an ablation study of Storyline Contextualizer for the story visualization task on the PororoSV.
As shown in Table \ref{tab:ablation_sc}, the proposed Storyline Contextualizer outperforms the other variants, achieving the lowest FID (13.61), highest Char. F1 score (77.24), and highest Frm. Acc. (51.59).
The Causal Storyline Contextualizer, which uses causal self-attention layers, limits each token embedding to only focus on previous token embeddings. This constraint prevents the storyline embedding from capturing the full context, leading to weaker performance compared to the proposed Storyline Contextualizer.
Overall, the results indicate that the proposed Storyline Contextualizer excels at maintaining consistency in characters and scenes across frames, outperforming the other variants.

\setlength{\tabcolsep}{8pt}
\begin{table}[t]
\scriptsize
\centering
\renewcommand\arraystretch{1}
    \begin{tabular}{lccc}
        \toprule
        \textbf{Number of Layers} & \textbf{FID $\downarrow$}            & \textbf{Char. F1 $\uparrow$}       & \textbf{Frm. Acc. $\uparrow$} \\
        \midrule
        1   & 13.96 & 75.20 & 48.26       \\
        2   & 13.84 & 76.06 & 49.56       \\  
        \textbf{4 (Ours)} & \textbf{13.61} & \textbf{77.24} & \textbf{51.59}  \\
        8   & 14.78	& 74.73	& 46.88       \\  
        \bottomrule
    \end{tabular}
  \caption{Ablation Study on the number of layers in the Storyline Contextualizer for the story visualization task on the PororoSV.}
  \label{tab:ablation_sc_num_layer}
\end{table}	

\setlength{\tabcolsep}{8pt}
\begin{table}[t]
  \scriptsize
  \centering
    \begin{tabular}{lccc}
        \toprule
        \textbf{Initialization Method} & \textbf{FID $\downarrow$}            & \textbf{Char. F1 $\uparrow$}       & \textbf{Frm. Acc. $\uparrow$} \\
        \midrule
        Random Initialization   & 13.84 & 76.06 & 49.56       \\  
        \textbf{Zero Initialization (Ours)}   & \textbf{13.61} & \textbf{77.24} & \textbf{51.59}       \\
        \bottomrule
    \end{tabular}
  \caption{Ablation study on initialization methods for the Storyline Contextualizer in the story visualization task on the PororoSV.}
  \label{tab:ablation_sc_init}
  \vspace{-5pt}
\end{table}

\setlength{\tabcolsep}{8pt}
\begin{table}[t]
\scriptsize
  \centering
    \begin{tabular}{lccc}
        \toprule
        \textbf{Initialization Method} & \textbf{FID $\downarrow$}            & \textbf{Char. F1 $\uparrow$}       & \textbf{Frm. Acc. $\uparrow$} \\
        \midrule
        Random Initialization   & 14.66	& 75.40 & 48.03 \\ 
        \textbf{Zero Initialization (Ours)}  & \textbf{13.61} & \textbf{77.24} & \textbf{51.59}       \\
        \bottomrule
    \end{tabular}
  \caption{Ablation study on initialization methods for the StoryFlow Adapter in the story visualization task on the PororoSV.}
  \label{tab:ablation_storyflow_init}
\end{table}

\setlength{\tabcolsep}{8pt}
\begin{table}[t]
\scriptsize
  \centering
    \begin{tabular}{lccc}
        \toprule
        \textbf{First Image Input Method} & \textbf{FID $\downarrow$}            & \textbf{Char. F1 $\uparrow$}       & \textbf{Frm. Acc. $\uparrow$} \\
        \midrule
        \textbf{Conv. First Image (Ours)}                  & \textbf{13.86} & \textbf{76.25}    & \textbf{50.72}     \\
        Concat. First Image                & 14.17          & 74.82             & 47.81             \\  \bottomrule
    \end{tabular}
  \caption{Ablation study of the first image input method for the story continuation task on the PororoSV.}
  \label{tab:first_image_input_method}
  \vspace{-5pt}
\end{table}

\noindent\textbf{Ablation study on the the number of layers in the Storyline Contextualizer.}
We conduct an ablation study on the the number of layers in the Storyline Contextualizer for the story visualization task on the PororoSV. 
As shown in Table \ref{tab:ablation_sc_num_layer}, the model's performance improves as the number of layers increases from 1 to 4. Specifically, the proposed 4-layer model achieves the best results, with the lowest FID (13.61), highest Char. F1 score (77.24), and highest Frm. Acc. (51.59). However, increasing the number of layers to 8 results in a decline in performance, with a higher FID (14.78) and lower Char. F1 (74.73) and Frm. Acc. (46.88). These results indicate that a 4-layer configuration provides the optimal balance between model complexity and performance, while additional layers may introduce unnecessary complexity that degrades the model's ability to maintain consistency and accuracy in the generated story frames.

\noindent\textbf{Ablation study on initialization methods for the Storyline Contextualizer.}
We conduct an ablation study on initialization methods for the Storyline Contextualizer in the story continuation task on the PororoSV.
As shown in Table \ref{tab:ablation_sc_init}, the model initialized with zero initialization achieves better results compared to random initialization. Specifically, zero initialization leads to a lower FID (13.61), higher Char. F1 score (77.24), and higher Frm. Acc. (51.59) than random initialization. These results indicate that zero initialization helps the model stabilize training and improves its ability to maintain consistency and accuracy in the generated story frames.

\noindent\textbf{Ablation study on initialization methods for the StoryFlow Adapter.}
We conduct an ablation study on initialization methods for the StoryFlow Adapter in the story continuation task on the PororoSV.
As shown in Table \ref{tab:ablation_storyflow_init}, the model initialized with zero initialization significantly outperforms the model initialized with random initialization. Specifically, zero initialization yields a lower FID (13.61), higher Char. F1 score (77.24), and higher Frm. Acc. (51.59) compared to random initialization, which results in an FID of 14.66, Char. F1 of 75.40, and Frm. Acc. of 48.03. These results suggest that zero initialization enhances the model's performance by providing more stable and effective training, leading to better consistency and accuracy in generating story frames.

\noindent\textbf{Ablation study of the first image input method in the story continuation.}
We conduct an ablation study on the input method of the first image of our ContextualStory in the story continuation task on the PororoSV in Table \ref{tab:first_image_input_method}.
\textit{Conv. First Image} means processing the first image using a convolution layer in each UNet block. \textit{Concat. First Image} means concatenating the latent representation of the first image with the latent noise of other images and then feeding it directly into UNet.
It can be observed that \textit{Conv. First Image} achieved the best results across all metrics, while \textit{Concat. First Image} showed a decline in performance. The superior performance of \textit{Conv. First Image} may be attributed to its incorporation of information from the first image into each UNet block, thus providing stronger guidance.

\section{Limitations}
One limitation of ContextualStory is its difficulty in maintaining optimal character layout and details when generating story frames with many characters.
This limitation primarily stems from the use of the pre-trained Stable Diffusion 2.1-base, meaning that the performance of ContextualStory is inherently constrained by the capabilities of Stable Diffusion. In future work, we plan to address these issues by leveraging more powerful text-to-image models, such as Stable Diffusion 3 \cite{esser2024scaling}, DALL-E 3 \cite{betker2023improving}, and PixArt-$\alpha$ \cite{chen2023pixartalpha}, to improve both character layout and detail in scenes with multiple characters. Additionally, we will consider incorporating layout control mechanisms to further improve character arrangement and detail in complex scenes.

\end{document}